\documentclass[lettersize,journal]{IEEEtran}

\usepackage{amsmath,amssymb,amsfonts}
\usepackage{algorithmic}
\usepackage{graphicx}
\usepackage{textcomp}
\usepackage{xcolor}
\usepackage{url}
\usepackage{makecell}
\usepackage{subcaption}
\usepackage{caption}
\usepackage{booktabs}
\usepackage{bbding}
\usepackage{array,multirow}
\usepackage{pgffor}
\usepackage{tikz}
\usepackage{makecell} 
\usepackage[font=small, textfont=sf,labelfont=rm, subrefformat=parens]{subcaption}

\usepackage[pagebackref=false,
            breaklinks=true,colorlinks,
            citecolor=cyan,linkcolor=red,
            bookmarks=false]{hyperref}
            
\newcommand{\firstone}[1]{\colorbox{red!15}{#1}}
\newcommand{\secondone}[1]{\colorbox{blue!15}{#1}}

\begin{document}
\title{ReviveDiff: A Universal Diffusion Model for Restoring Images in Adverse Weather Conditions}

\author{Wenfeng Huang, 
        Guoan Xu, 
        Wenjing Jia,~\IEEEmembership{Member,~IEEE,} 
        Stuart Perry,
       and Guangwei Gao,~\IEEEmembership{Senior Member,~IEEE}

\thanks{This work was supported in part by the foundation of Key Laboratory of Artificial Intelligence of Ministry of Education under Grant AI202404.~\textit{(Corresponding author: Wenjing Jia; Guangwei Gao.)}}   

\thanks{Wenfeng Huang, Guoan Xu, Wenjing Jia, and Stuart Perry are with the Faculty of Engineering and Information Technology, University of Technology Sydney, Sydney, NSW 2007, Australia (e-mail: huang-wenfeng@outlook.com, xga\_njupt@163.com, Wenjing.Jia@uts.edu.au, Stuart.Perry@uts.edu.au).}

\thanks{Guangwei Gao is with the PCA Lab, Key Lab of Intelligent Perception and Systems for High-Dimensional Information of Ministry of Education, School of Computer Science and Engineering, Nanjing University of Science and Technology, Nanjing 210094, China, and also with the Key Laboratory of Artificial Intelligence, Ministry of Education, Shanghai 200240, China (e-mail: csggao@gmail.com).}

}

\markboth{IEEE Transactions on Image Processing}%
{Shell \MakeLowercase{\textit{et al.}}: A Sample Article Using IEEEtran.cls for IEEE Journals}

\maketitle

\begin{abstract}Images captured in challenging environments--such as nighttime, smoke, rainy weather, and underwater--often suffer from significant degradation, resulting in a substantial loss of visual quality. 
The effective restoration of these degraded images is critical for the subsequent vision tasks. 
While many existing approaches have successfully incorporated specific priors for individual tasks, these tailored solutions limit their applicability to other degradations. 
In this work, we propose a universal network architecture, dubbed ``ReviveDiff'', which can address various degradations and restore images to their original quality by enhancing and restoring their details.  
Our approach is inspired by the observation that, unlike degradation caused by movement or electronic issues, quality degradation under adverse conditions primarily stems from natural media (such as fog, water, and low luminance), which generally preserves the original structures of objects. 
To restore the quality of such images, we leveraged the latest advancements in diffusion models and developed ReviveDiff to restore image quality from both macro and micro levels across some key factors determining image quality, such as sharpness, distortion, noise level, dynamic range, and color accuracy. We rigorously evaluated ReviveDiff on seven benchmark datasets covering five types of degrading conditions: Rainy, Underwater, Low-light, Smoke, and Nighttime Hazy. Our experimental results demonstrate that ReviveDiff outperforms the state-of-the-art methods both quantitatively and visually.

\end{abstract}
\begin{IEEEkeywords}
Image Restoration, Diffusion Model, Adverse Conditions.
\end{IEEEkeywords}

\section{Introduction}

\label{sec:introduction}

\IEEEPARstart{I}{mages} captured in adverse environments, such as rain, fog, underwater conditions, and low-light scenarios, frequently endure significant degradation in quality. 
These natural factors disrupt light propagation, leading to substantial losses in visual clarity, color fidelity, and overall image visibility and usability. 
Furthermore, as these challenging conditions often coincide with low illumination,  the need for effective enhancement and restoration of image quality is paramount to ensuring that the subsequent vision tasks can be performed accurately and reliably. 

Over the years, substantial progress has been made in addressing various image degradation challenges, including 
denoising~\cite{liu2024residual}, 
dedarkening~\cite{retinexnet}, and deraining~\cite{DerainNet}. 
These methods typically rely on specific priors tailored to individual tasks. 
For instance, Retinex theory~\cite{land1977retinex} has been successfully employed in many low-light enhancement techniques~\cite{retinexnet}, 
while the atmospheric scattering model~\cite{gta5} has been commonly applied to guide the dehazing efforts. 
Fourier~\cite{UHDFour} and wavelet transformations have been extensively employed to address issues such as low resolution, blurring, and noise. 
However, these existing approaches are largely specialized, focusing on singular tasks with highly tailored solutions. This specialization, while effective within narrow scopes, limits its applicability across different types of degradation. 

\begin{figure}[t]
	\centering  
	\includegraphics[width = 1\linewidth]{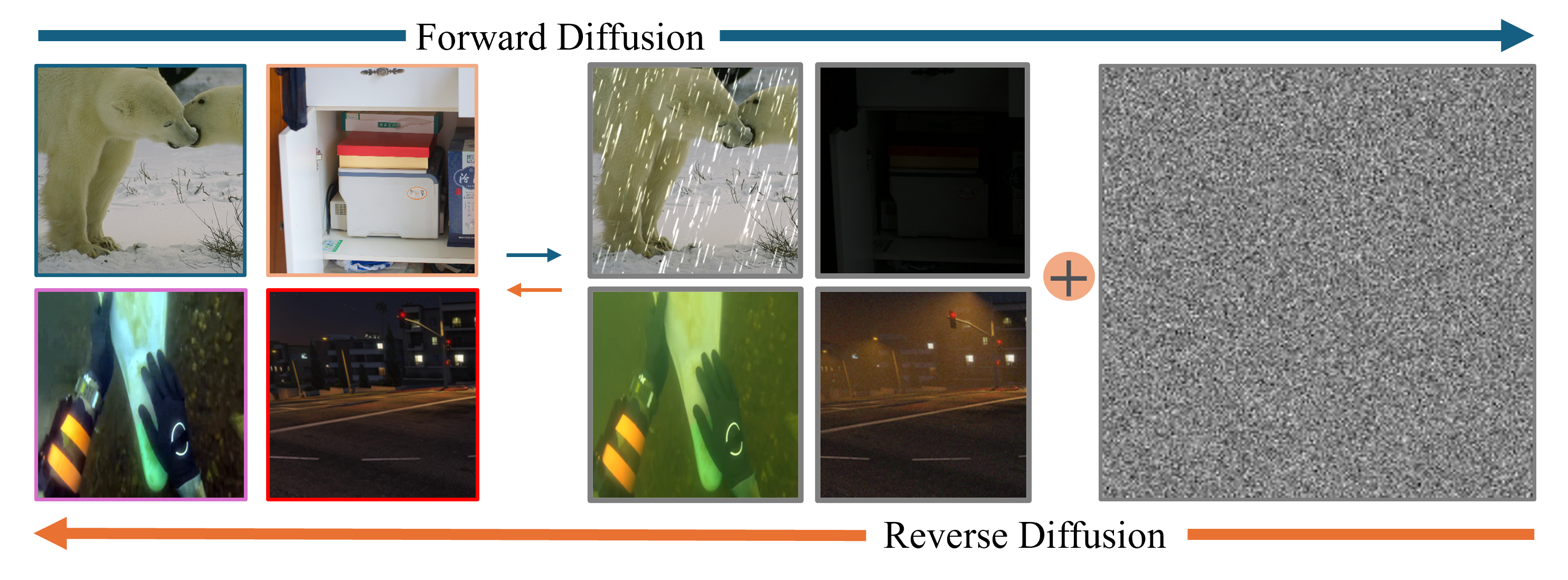}
        \vspace{-1em}
        \caption{Our ReviveDiff can address various adverse degradations (such as rain, low light, underwater, and nighttime dehazing) by enhancing and restoring their quality.}
	\vspace{-2em}
	\label{diff}
\end{figure}

As research has progressed, more complex challenges, such as nighttime image dehazing~\cite{gta5,yan2020nighttime} and underwater image enhancement~\cite{utrans,fivenet,NU2Net,du2023uiedp}, have emerged. 
Although these solutions have achieved impressive results by employing techniques such as Atmospheric Point Spread Function-guided Glow Rendering and posterior distribution processes for underwater color restoration, they remain constrained by their focus on specific degradation scenarios. 
Their tailored design potentially limits their applicability to other degradations. 

Recognizing the limitations of task-specific models, recent efforts such as NAFNet~\cite{nafnet}, IR-SDE~\cite{IRSDE}, and Restormer~\cite{restormer} have aimed to address common degradations with a single neural network architecture, dealing with issues such as blurring, noise, and low resolution. 
However, these methods often fall short in handling images from particularly challenging environments, such as those captured at night or underwater, which underscores the need for a universal framework capable of concurrently addressing diverse adverse weather conditions.

The root causes of image degradation vary significantly; for example, blurring is typically caused by motion, while noise and low resolution often stem from limitations in imaging equipment. 
In contrast, quality degradation under adverse conditions is predominantly due to natural phenomena such as fog, water, and low light. 
This type of degradation generally maintains the structural integrity of objects in the image, leading to less distortion and information loss than other degradation types. 
These characteristics have inspired us to leverage general image attributes, such as edges, noise levels, and color distortions, to guide the distortion process across a wide range of adverse weather conditions, rather than relying on priors specific to an individual task. 

To address this need for a more versatile and robust solution, we propose ReviveDiff, a groundbreaking universal diffusion model designed to restore image quality across a wide range of adverse environments. 
Our approach builds on the strengths of diffusion models, particularly their use of Stochastic Differential Equations (SDE) and Gaussian noise-based features, to create a model capable of handling an expansive feature space with exceptional learning capabilities. 
Unlike existing models, ReviveDiff is not confined to a single type of degradation but is equipped to tackle a wide range of adverse conditions and restore image qualities. 

Specifically, what sets our ReviveDiff apart is its ability to address key factors that determine image quality, such as sharpness, contrast, noise, color accuracy, and distortion, from both a coarse-level overview and a detailed, fine-level perspective. 
Our model incorporates a novel multi-stage Multi-Attentional Feature Complementation module, which integrates spatial, channel, and point-wise attention mechanisms with dynamic weighting to balance macro and micro information. This design significantly enhances information integration, ensuring that the restored images are not only visually appealing but also retain the structural and color integrity necessary for subsequent tasks.
Our comprehensive experiments, conducted across seven benchmark datasets and five types of adverse conditions—Rainy, Underwater, Low-light, Foggy, and Nighttime hazy—demonstrate that ReviveDiff consistently outperforms state-of-the-art (SOTA) methods 
in both quantitative measures and visual quality. This highlights the model’s broad applicability and its potential to set new standards in image restoration under challenging conditions.

The main contributions of our work are listed as follows:
\begin{enumerate}

\item We developed ReviveDiff, a universal framework that can restore image quality across a wide range of adverse weather conditions, representing a significant advancement over the existing task-specific models.

\item We proposed a novel Coarse-to-Fine Learning Block (C2FBlock), which not only expands the receptive field through two distinct scale branches 
but also minimizes information loss. This design enables the network to capture complex and challenging degradations effectively.

\item We designed a novel Multi-Attentional Feature Complementation (MAFC) module that integrates spatial, channel, and pixel attention mechanisms with dynamic weighting. This helps the model effectively complement information between macro and micro levels.

\item We introduced a unique prior-guided loss function that ensures optimal pixel restoration by leveraging edge information to refine object shapes and structures, while utilizing histogram information to guide accurate color correction under adverse conditions.

\end{enumerate}
\vspace{2mm}

\begin{figure*}[!h]
	\centering
	\includegraphics[width = 1\linewidth]{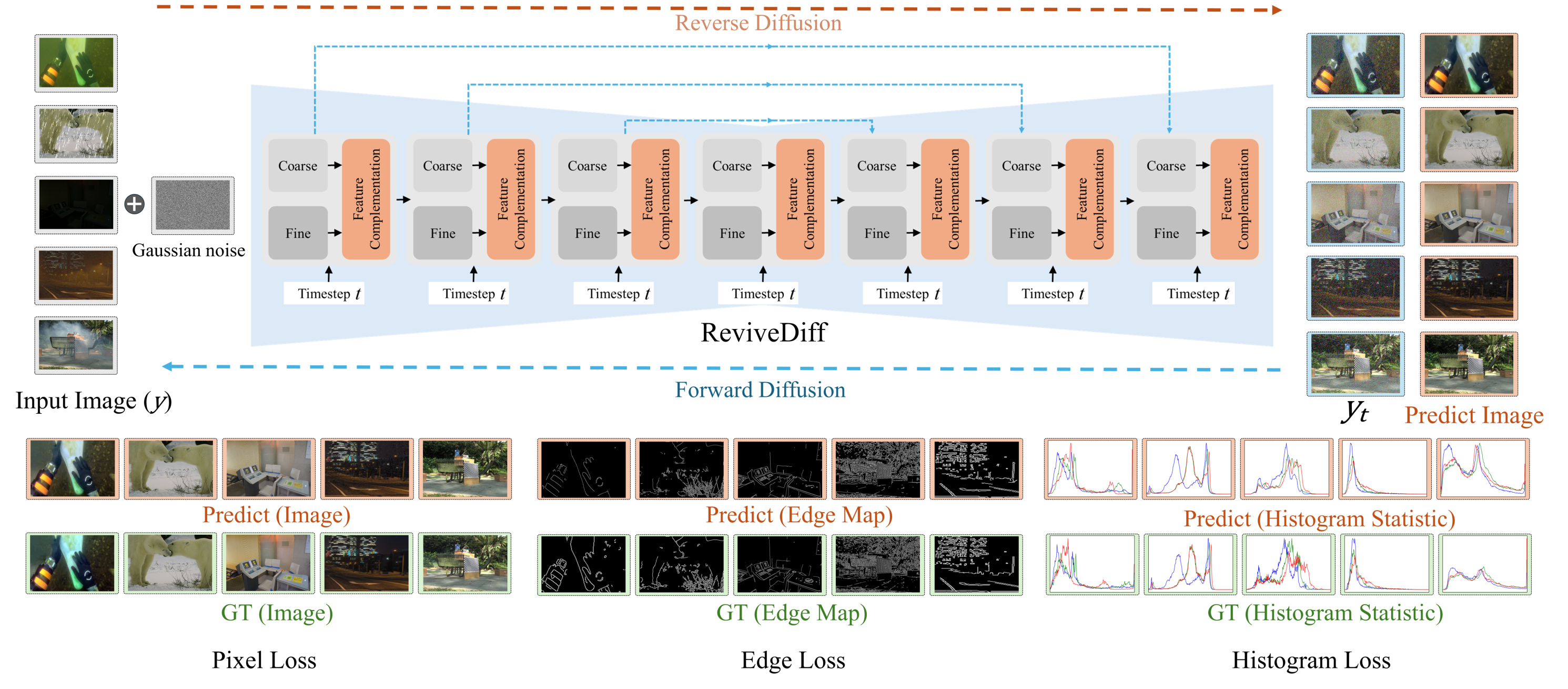}
	\vspace{-1em}
    \caption{The basic framework of our proposed ReviveDiff, which is a U-shaped latent diffusion model with stacked Coarse-to-Fine Learning Blocks (C2FBlock) and Multi-Attentional Feature Complementation (MAFC) modules, guided by a combined loss to tackle diverse real-world image restoration tasks.}
     \vspace{-1em}
	\label{fig:overall}        
\end{figure*}


\section{Related Work}
\label{sec:Related Work}

In this section, we briefly review recent advancements in image restoration, particularly in challenging environments, and discuss the role of diffusion models in vision tasks. 

\subsection{Image Restoration}

Image restoration is a broad field aimed at recovering high-quality images from degraded inputs. 
Numerous single-task approaches have been proposed to tackle specific problems, such as deblurring, denoising, deraining, and super-resolution. 
For instance, MDARNet~\cite{MDARNet} employs dual attention for image deraining, and PREnet~\cite{PREnet} focuses on progressive rain removal. 
In the context of dark conditions, 
Li \textit{et al.}~\cite{LLformer} introduced LLFormer with the Fourier Transform for low-light restoration. 
Underwater image restoration includes MLLE~\cite{mlle}, which uses locally adaptive contrast, and UTrans~\cite{utrans}, which integrates multi-scale feature fusion. 
For nighttime enhancement, Jin \textit{et al.}~\cite{gta5} developed a gradient-adaptive convolution
, and Yan \textit{et al.}~\cite{yan2020nighttime} used high-low frequency decomposition. 
Although these methods demonstrate strong performance, each primarily addresses a single type of degradation or environment, limiting their applicability across multiple adverse conditions.

In contrast, multi-task approaches aim to generalize or unify restoration across various tasks. 
SFNet~\cite{SFNet} employed a multi-branch module with frequency information for broader image restoration, while Uformer~\cite{Uformer} leveraged a U-shaped Transformer for multiple tasks. 
Restormer~\cite{restormer} was designed for high-resolution restoration, while MAXIM~\cite{MAXIM} introduced a multi-axis MLP structure to handle diverse degradations. 
Liu \textit{et al.}~\cite{PROMPTGIP} proposed PromptGIP, which adopts a visual prompting and question-answering paradigm to unify various image processing tasks. 
Zheng \textit{et al.}~\cite{DIFFUIR} presented a diffusion-based universal restoration model with selective hourglass mapping, and Luo \textit{et al.}~\cite{DACLIP} combined vision-language models for multi-task image restoration. 
Liu \textit{et al.}~\cite{DIFFPlugin} introduced a lightweight Task-Plugin module featuring a dual-branch architecture within a diffusion model to supply task-specific priors for low-level tasks.
Sun \textit{et al.}~\cite{sun2024restoring} proposed Histoformer, with a dubbed histogram self-attention mechanism for image restoration.
By accommodating different degradation types, these multi-task solutions offer broader applicability, though balancing performance across diverse conditions remains an ongoing challenge.

\subsection{Diffusion Models in Vision}

The diffusion models~\cite{ddim} have emerged as powerful tools in the field of image generation and restoration and have achieved significant improvement.
These models work by degrading a signal through Gaussian noise and then restoring it through a reverse process. 
Fei \textit{et al.}~\cite{gdp} introduced a unified image restoration model that utilizes a diffusion prior.
In another significant contribution, Luo \textit{et al.} proposed IR-SDE~\cite{IRSDE}, which focuses on image restoration through mean-reverting stochastic differential equations. 

The application of diffusion models has also extended to enhancing images captured in adverse conditions. 
For example, Jiang \textit{et al.} proposed LightenDiffusion~\cite{LightenDiffusion}, a latent-Retinex-based diffusion model specially designed for low-light image enhancement. 
For underwater image enhancement, 
Du \textit{et al.} proposed UIEDP~\cite{du2023uiedp}, which employs a diffusion prior tailored for underwater images, and Tang \textit{et al.}~\cite{dmuie} introduced a transformer-based diffusion model. Zheng \textit{et al.}~\cite{DIFFUIR} proposed a universal image restoration framework based on a diffusion model that leverages a selective hourglass mapping strategy. Wang \textit{et al.}~\cite{wang2025reconciling} introduced RDMD, a framework that leverages a single pre-trained diffusion model to derive two complementary regularizers. Kim \textit{et al.}~\cite{kim2024rad} proposed region-aware diffusion models for image inpainting. DIFFPIR~\cite{DIFFPlugin}, a lightweight task-plugin module introduced by Liu \textit{et al.}, features a dual-branch architecture within a diffusion model that provides task-specific priors for low-level tasks.
The flexibility of diffusion models in accommodating different scenarios in these areas has inspired us to leverage the powerful generalization capabilities of diffusion models to build a universal solution--ReviveDiff-- that can effectively address a wide range of adverse degradations. 
Furthermore, to further enhance its effectiveness in tackling diverse real-world image restoration tasks, we adopted the Mean-Reverting Stochastic Differential Equations (SDEs) based diffusion models~\cite{IRSDE,luo2023refusion}, which implicitly model the degradation and apply to diverse tasks without changing the architecture. 

\section{Methodology}
\label{sec:Methodology}

As illustrated in Fig.~\ref{fig:overall}, our ReviveDiff is a U-shaped latent diffusion model, with stacked Coarse-to-Fine Blocks (C2FBlocks) and Multi-Attentional Feature Complementation (MAFC) modules to tackle diverse real-world image restoration tasks. 
Compared with other diffusion models used for image enhancement, our ReviveDiff performs fusion at both low-resolution, macro-level latent space and high-resolution, micro-level latent space from the original input for the decoding process, and under the guidance of fine granularity.  

\begin{figure}[!t]
	\centering
	\includegraphics[width = 1\linewidth]{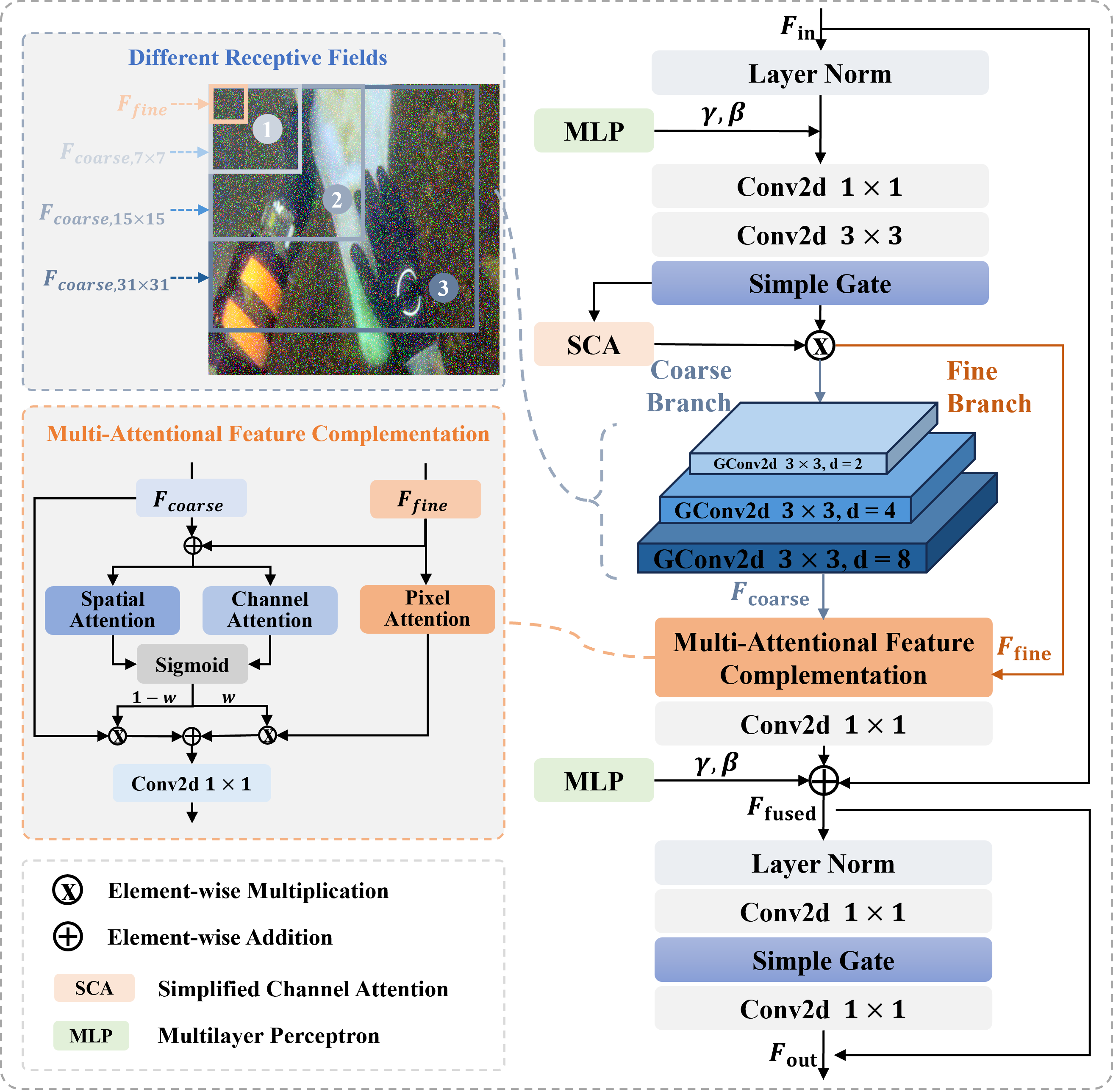}
	\caption{The Coarse-to-Fine Learning Block (C2FBlock) at the core of our proposed ReviveDiff network.}
  \vspace{-1em}
	\label{fig:dualbranch}
\end{figure}

Building on the insights from the previous research~\cite{IRSDE,luo2023refusion,nafnet}, the C2FBlock introduces a dual-branch structure purposefully designed to capture features at varying levels of granularity. 
Specifically, as shown in Fig.~\ref{fig:dualbranch}, the Coarse Branch allows the fusion to be performed with a larger receptive field (31 $\times$ 31), enabling it to capture broader contextual information, while the Fine Branch utilizes a focused receptive field of $3 \times 3$ to capture finer, more detailed features. 

\begin{figure*}[t]
	\centering
	\includegraphics[width = 1\linewidth]{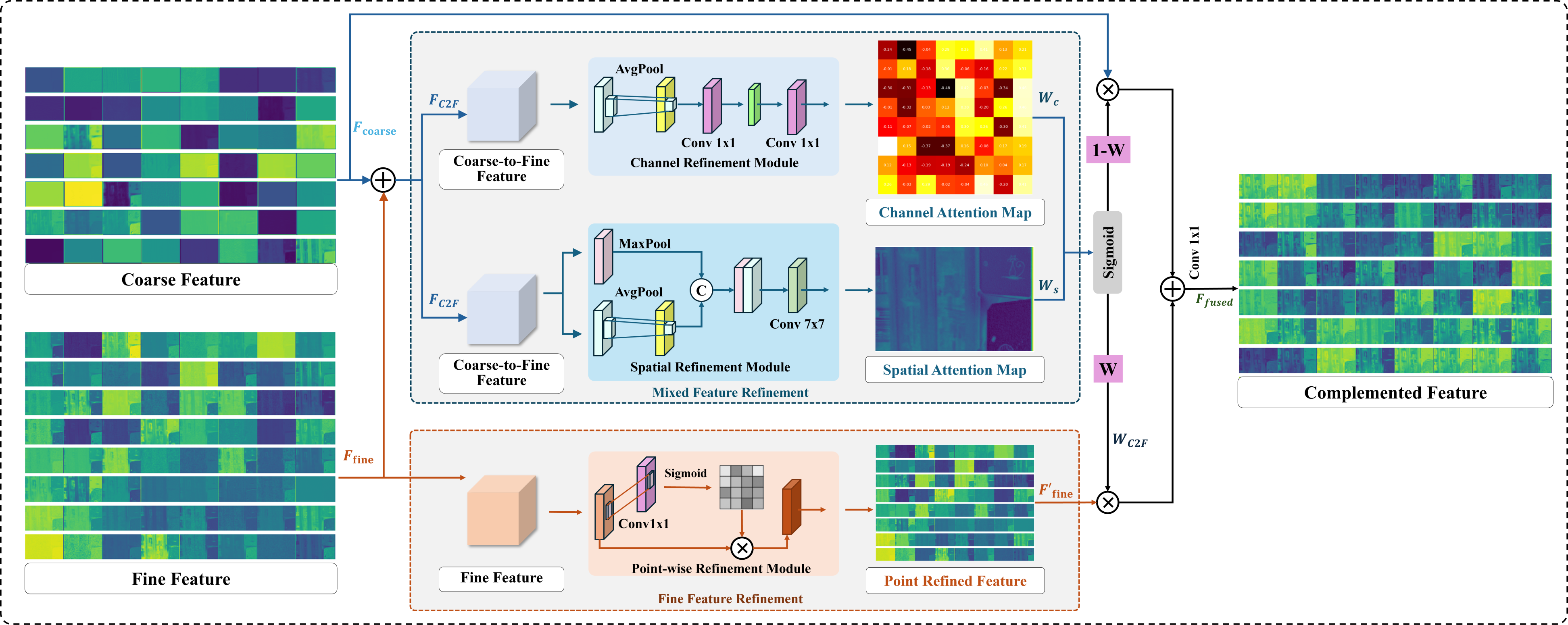}
	\caption{Our Multi-Attentional Feature Complementation (MAFC) module adaptively integrates coarse and fine features through complementary refinement pathways. The coarse stream undergoes dual-domain enhancement via channel-wise recalibration and spatial context aggregation, while a point-wise feature refinement module refines the fine branch. The two branches are then fused through an adaptive weighting strategy, resulting in complemented feature maps that preserve both global context and local details.}
  \vspace{-1em}
	\label{fig:attention}
\end{figure*}

Then, to effectively integrate these coarse and fine features, we developed a Multi-Attentional Feature Complementation module (see Fig.~\ref{fig:attention}), 
which incorporates three distinct types of attention mechanisms and employs dynamic weighting to adjust the balance between the contributions of Coarse and Fine features. 
This ensures an optimal balance that enhances the model's ability to restore image quality across diverse scenarios accurately.

\subsection{Mean-Reverting Stochastic Differential Processes}

In this work, we leverage a probabilistic diffusion approach to enhance visibility in low-light images. Specifically, our approach is based on a score-based generative framework that utilizes Mean-Reverting SDE~\cite{IRSDE} as the base diffusion framework to model the image-reviving diffusion process. 
Fig.~\ref{fig:overall} illustrates this process. 
The forward SDE gradually transforms the initial data distribution \(y_0\) into a Gaussian noise representation \(y_T\) over \(T\) steps. 
The objective of the reverse SDE process is then to reconstruct a high-quality image from this noisy representation \(y_T\).

\textbf{Forward SDE:} 
The forward SDE process is defined as the transformation of an input high-quality image into Gaussian noise. 
Mathematically, this process is expressed as
\begin{equation}
    \label{eq:forward_sde}
    dy = \alpha(t)\left(\nu - y\right)dt + \beta(t)dV,
\end{equation}
where \(\alpha(t)\) regulates the speed of mean reversion, \(\beta(t)\) denotes the level of stochastic volatility, and \(V\) is the standard Wiener process. 

To achieve a solvable form of the forward SDE, 
we set the condition \( \frac{\beta(t)^2}{\alpha(t)} = 2\kappa^2 \), with \(\kappa^2\) symbolizing the stationary variance for any $t\in[0, T]$. 
Assuming \(y(s)\) depicts the noisy image derived from high-quality input at time $s<t$, the solution to Eq.~\ref{eq:forward_sde} is given by~\cite{IRSDE}:
\begin{equation}
    y(t) = \nu + (y(s) - \nu)e^{-\hat{\alpha}(s:t)} + \int_s^t \beta(u)e^{-\hat{\alpha}(u:t)}dV(u),
\end{equation}
where \(\hat{\alpha}(s:t) = \int_s^t \alpha(u)du\). The distribution of \(y_t\) is determined as follows:
\begin{equation}
y_t \sim p_t(y) = \mathcal{N}(y_t | \mu_t(y), \sigma_t^2),
\end{equation}
where 
\begin{equation}
\mu_t(y) = \mu + (y(s) - \mu) \mathrm{e}^{-\hat{\alpha}(s:t)},
\end{equation}
and 
\begin{equation}
\sigma_t^2 = \kappa^2 \left(1 - \mathrm{e}^{-2 \hat{\alpha}(s:t)}\right).
\end{equation}


\textbf{Reverse SDE:} 
The goal of the Reverse SDE process is to reconstruct the high-quality image from its degraded representation \(y(T)\). 
As established in earlier research, the reverse SDE is formulated as
\begin{equation}
dy = \left[\alpha(t)(\nu - y) - \beta(t)^2\nabla_y\log p_t(y)\right]dt + \beta(t)d\tilde{V},
\end{equation}
where \(\tilde{V}\) signifies another instance of the Wiener process. 

The training mechanism utilizes high-quality images to enable the calculation of the gradient score:
\begin{equation}
\nabla_y\log p_t(y) = -\frac{y(t) - \mu_t(y)}{\sigma_t^2},
\end{equation}
facilitating the sampling of \(y(t)\) as \(y(t) = \mu_t(y) + \sigma_t \zeta_t\), with \(\zeta_{t} \sim \mathcal{N}(0, I)\) indicating Gaussian noise. 

Thus, the score function is derived as
\begin{equation}
\nabla_y\log p_t(y) = -\frac{\zeta_t}{\sigma_t}.
\end{equation}


\subsection{Coarse-to-Fine Learning}

Unlike the existing diffusion models used for image enhancement, our ReviveDiff performs feature fusion in the latent space at both macro and micro levels and fuses them under the guidance of fine granularity.  
This is achieved through a series of stacked C2FBlocks at the U-Net architecture's core. 

As illustrated in Fig.~\ref{fig:dualbranch}, our C2FBlock is specially designed to capture features at two distinct levels of granularity: The Fine Branch is responsible for capturing the image's detailed, localized features with a small receptive field, 
whereas the Coarse Branch operates on a significantly larger receptive field, focusing on the image's global contextual information in low-resolution space. 
This dual-branch approach ensures that both macro-level contextual information and detailed, pixel-level details are effectively captured and integrated. 

In addition, to improve computational efficiency, we adopt the Simple Gate (SC) activation~\cite{nafnet} to replace more complex nonlinear activation functions. The Simple Gate achieves the effect of nonlinear activation using a single multiplication operation, which is especially advantageous in preserving critical information in regions of low pixel values.

Specifically, let  $\mathcal{F}_{in}$ represent the input feature. The features extracted by the Fine Branch can be represented as
\begin{equation}
\mathcal{F'}_{fine} = DWConv(LayerNorm(\alpha_{1} \odot \mathcal{F}_{in} + \beta{1})).
\end{equation}
The Fine Branch uses a $3 \times 3$ receptive field to capture detailed local features, producing $\mathcal{F'}_{fine}$. 
Then, the final output of the Fine Branch $\mathcal{F}_{fine}$ is computed using SimpleGate as $SG$ and a Simplified Channel Attention as $SCA$, as
\begin{equation}
\mathcal{F}_{fine} = SCA(SG(\mathcal{F'}_{fine})).
\end{equation}

In contrast to the Fine Branch, the Coarse Branch focuses on global information and operates on a significantly larger receptive field of $31 \times 31$, as depicted in Fig.~\ref{fig:attention}.
However, existing approaches, such as the non-local mechanism, vision transformer blocks, and convolutional layers with large kernel sizes, tend to be computationally expensive, consuming significant resources and slowing down processing speeds.  
To address this issue, we utilize grouped-dilated convolutions to efficiently expand the receptive field size without substantially increasing the computation load. 

To mitigate the information loss due to the gridding effect commonly associated with dilated convolutions, we employ a stacking strategy rather than a single convolutional layer. 
Specifically, we apply increasing dilation rates of 2-4-8 to achieve a large receptive field while minimizing the loss of fine-grained details. 
Finally, grouped convolutions are used to prevent a substantial increase in parameters, ensuring that the model remains computationally efficient. 

Thus, denoting grouped-dilated convolution with a dilation rate of $i$ as $f_{r=i}$, the Coarse Feature, $\mathcal{F}_{coarse, size}$, where $size$ represents its kernel size, can be obtained as
\begin{equation}
\left.\left\{\begin{array}{ll}
\mathcal{F}_{coarse,7\times7} = f_{r=2}(\mathcal{F}_{fine})\\
\mathcal{F}_{coarse,15\times15} = f_{r=4}(\mathcal{F}_{coarse,7\times7})\\
\mathcal{F}_{coarse,31\times31} = f_{r=8}(\mathcal{F}_{coarse,15\times15})
\end{array}\right.\right..
\end{equation}
This coarse-to-fine learning process enables our model to capture both broad contextual information and focused local details, ensuring a balanced and thorough restoration of images across different types of degradations. 
As experimental results show, by fusing coarse and fine features in the latent space, our ReviveDiff achieves superior performance in a range of challenging image enhancement tasks.

\begin{figure*}[h]
    \centering
    \captionsetup[subfigure]{labelformat=empty, font=footnotesize}
    \setcounter{subfigure}{0}

    \subfloat[Input]{\includegraphics[width=0.17\textwidth]{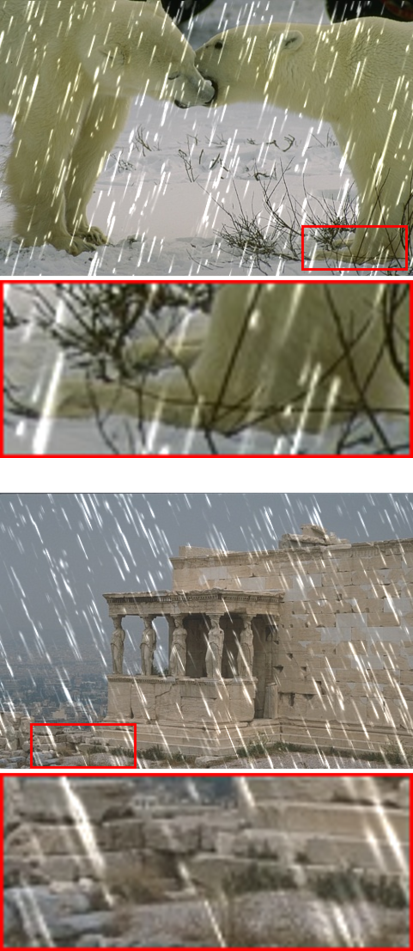}\vspace{-1mm}}\hfill
    \subfloat[DIDMDN~\cite{DIDMDN}]{\includegraphics[width=0.17\textwidth]{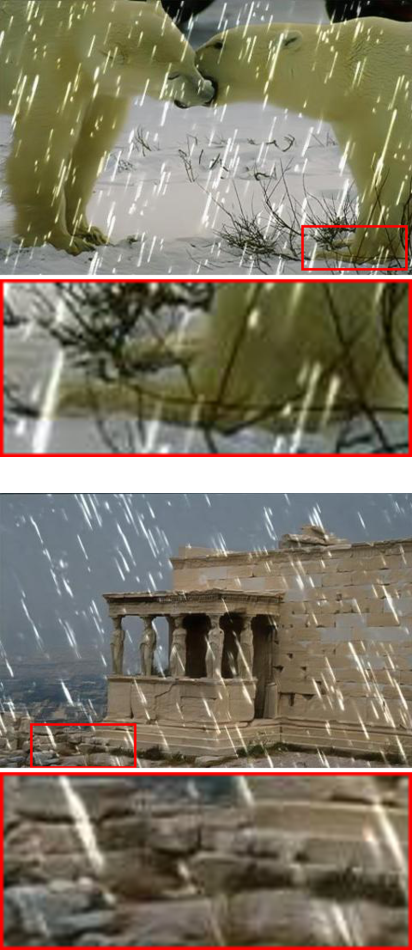}\vspace{-1mm}}\hfill
    \subfloat[DerainNet~\cite{DerainNet}]{\includegraphics[width=0.17\textwidth]{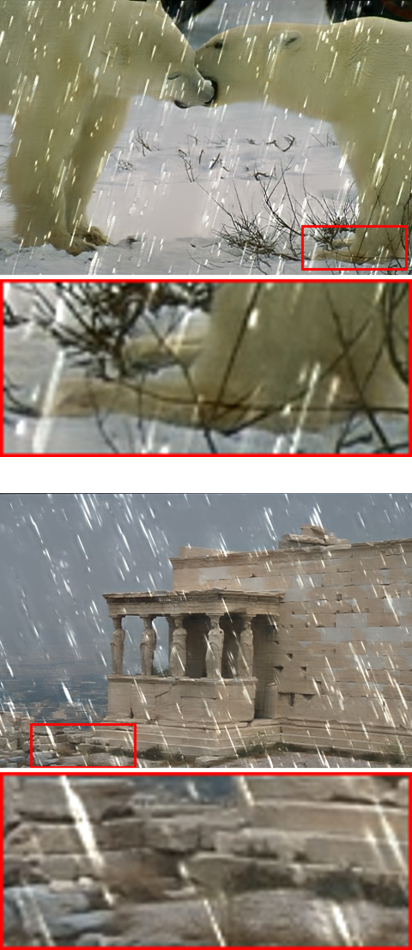}\vspace{-1mm}}\hfill
    \subfloat[SEMI~\cite{semi}]{\includegraphics[width=0.17\textwidth]{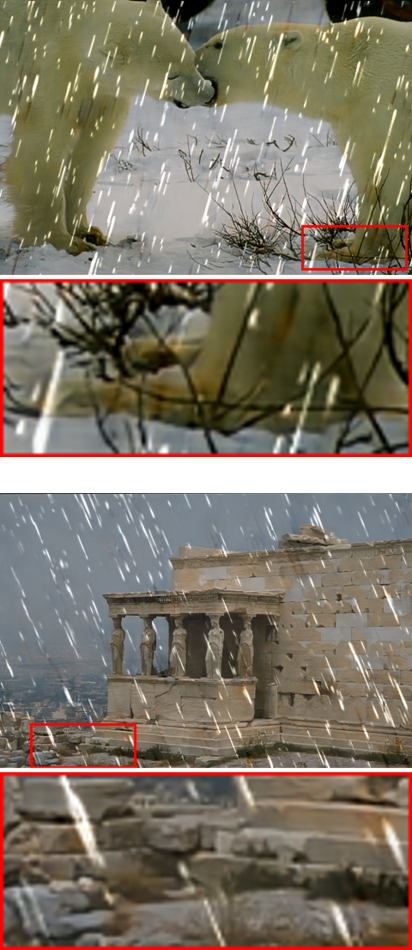}\vspace{-1mm}}\hfill
    \subfloat[UMRL~\cite{UMRL}]{\includegraphics[width=0.17\textwidth]{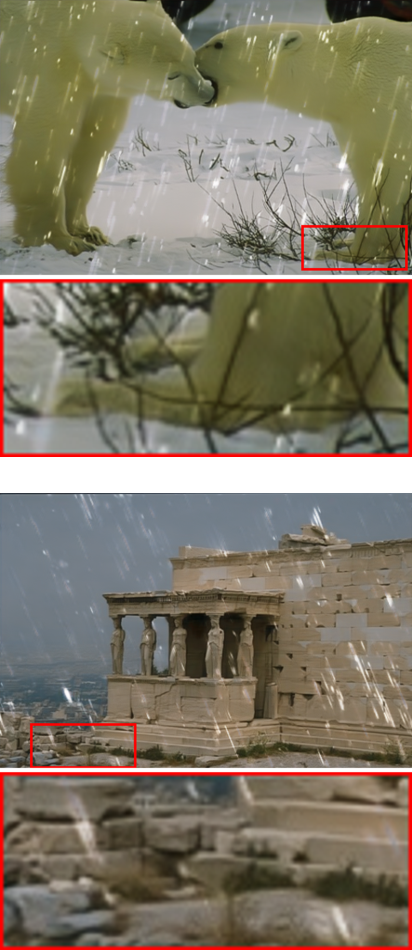}\vspace{-1mm}}\hfill \par\vspace{1mm}
    \subfloat[MPRNET~\cite{Zamir2021MPRNet}]{\includegraphics[width=0.17\textwidth]{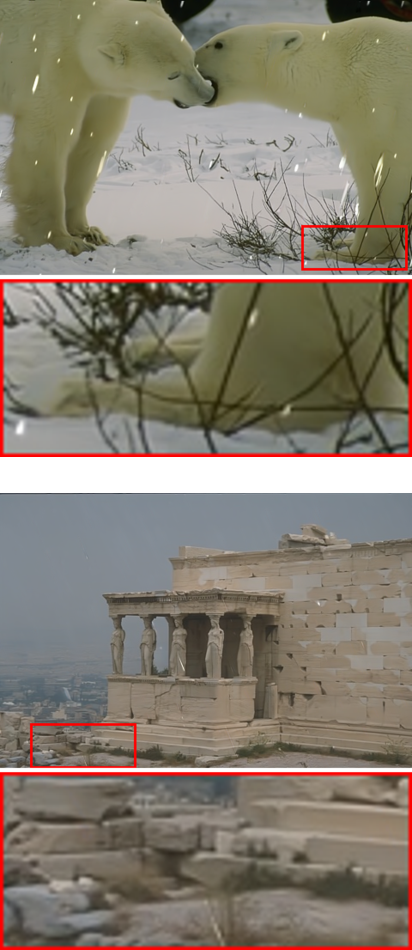}\vspace{-1mm}}\hfill
    \subfloat[MAXIM~\cite{MAXIM}]{\includegraphics[width=0.17\textwidth]{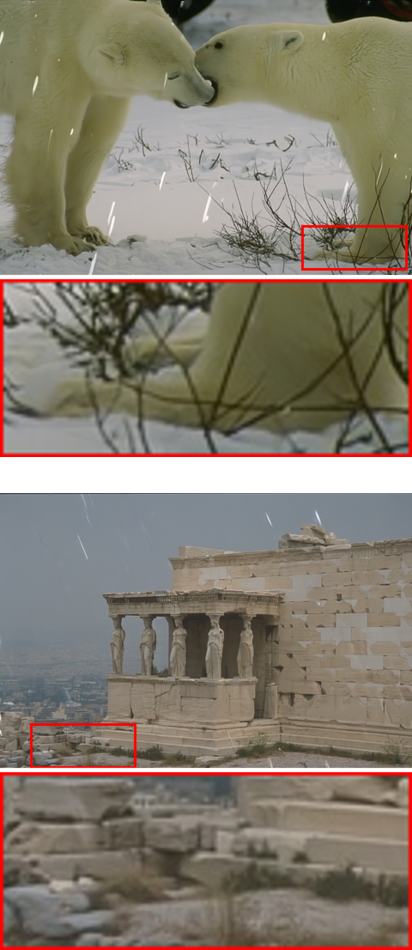}\vspace{-1mm}}\hfill
    \subfloat[IR-SDE~\cite{IRSDE}]{\includegraphics[width=0.17\textwidth]{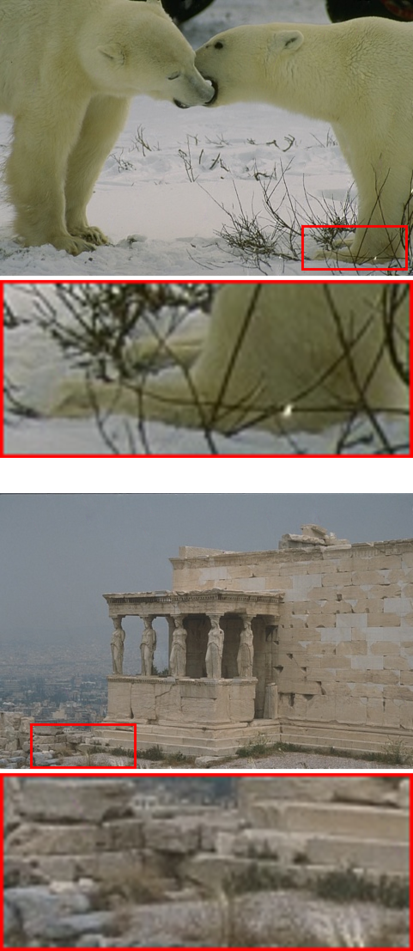}\vspace{-1mm}}\hfill
    \subfloat[\textbf{ReviveDiff (Ours)}]{\includegraphics[width=0.17\textwidth]{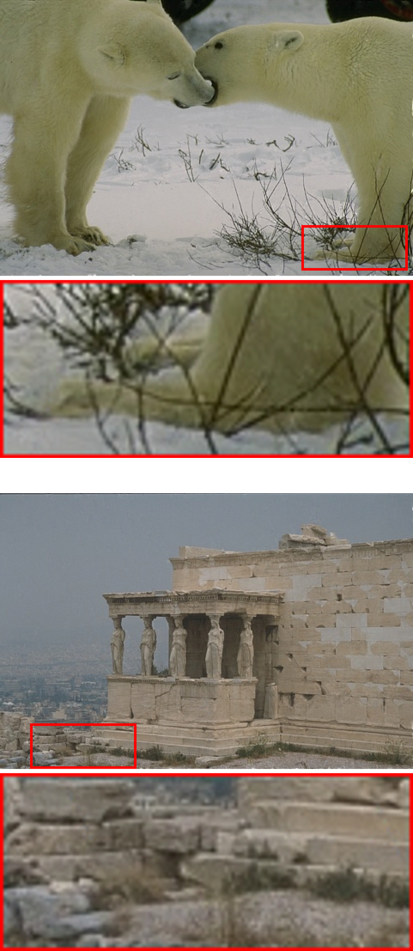}\vspace{-1mm}}\hfill
    \subfloat[GT]{\includegraphics[width=0.17\textwidth]{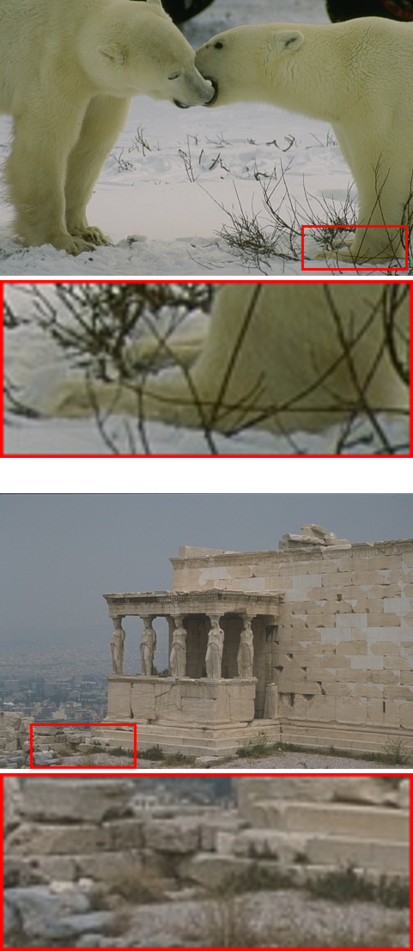}\vspace{-1mm}}\hfill \par\vspace{1mm}
    \setcounter{subfigure}{0}

    \caption{Visual comparison of the deraining results obtained with our ReviveDiff and SOTA approaches on Rain100L~\cite{Rain100}. The small images enclosed in red below each result provide an enlarged view of the areas highlighted by red rectangles.}
    \label{Rain100L:fig}
\end{figure*}

\subsection{Multi-Attentional Feature Complementation}
As illustrated in Fig.~\ref{fig:attention}, the proposed Multi-Attentional Feature Complementation (MAFC) module adaptively integrates coarse and fine features via complementary refinement branches. Specifically, coarse features undergo channel and spatial refinements, while fine features are refined by a point-wise feature refinement module. The two branches are then fused through an adaptive weighting strategy, resulting in complemented feature maps that preserve both global context and local details.

Given the Fine Features $\mathcal{F}_{fine}$ and the Coarse Features $\mathcal{F}_{coarse}$ obtained from Coarse-to-Fine Learning, the significant granularity gap between them poses a challenge for effective fusion and refinement. 

To address this issue, we propose a Multi-Attentional Feature Complementation (MAFC) module, 
which utilizes attention-based mechanisms~\cite{attfusion,chen2023dea} to dynamically compute feature weights both spatially~\cite{CBAM} and across channels~\cite{channelatt}. 
This allows for realignment and complementary fusion of the Coarse and Fine features, ensuring that the resulting feature map maintains critical information from both levels of granularity adaptively and complementarily. 

For spatial dimension alignment, we utilize a spatial attention mechanism to learn the spatial weight map $W_{\text{s}} \in \mathbb{R}^{H \times W}$. This map highlights the importance of different regions within an image.
Meanwhile, the channel weight is computed as $W_{\text{c}} \in \mathbb{R}^{C \times 1 \times 1}$, indicating the importance of each channel for a given task.

Denote $\mathcal{F}_{C2F} = \mathcal{F}_{coarse} + \mathcal{F}_{fine}$ as the feature map integrating both Coarse and Fine Features,  
the spatial weight $W_{\text{s}}$ and the channel weight $W_{\text{c}}$ can be calculated as 
\begin{equation}
    \left.\left\{\begin{array}{ll}
    W_{s}&={Cov}_{7\times7}([GAP_s(\mathcal{F}_{C2F}),GMP_s(\mathcal{F}_{C2F})]),\\
    W_{c}&={Conv}_{1\times1}(ReLU({Conv}_{1\times1}(GAP_c(\mathcal{F}_{C2F}))).
    \end{array}\right.\right.
\end{equation}
Here, $GAP$ refers to Global Average Pooling, $GMP$ denotes Global Max Pooling, and $s$ and $c$ indicate that the GAP/GMP operation is conducted along the spatial or channel dimensions, respectively. 

Thus, the dynamic combination weight \(W_{C2F}\) recalibrating \(\mathcal{F}_{coarse}\) and \(\mathcal{F}_{fine}\) can be defined as
\begin{equation}
W_{C2F} = \sigma(W_{s} + W_{c}),
\end{equation}
where \(\sigma\) denotes the Sigmoid function.

Furthermore, to further refine the fine features \(\mathcal{F}_{fine}\), we employ a Pixel Attention~\cite{pixatt} scheme to establish attention coefficients at the pixel level, aiming to obtain a more sophisticated feature map \(\mathcal{F'}_{fine}\):
\begin{equation}
\mathcal{F'}_{fine} = \sigma(Conv_{1\times1}(\mathcal{F}_{fine})) \cdot \mathcal{F}_{fine}.
\end{equation}

Finally, the feature map \(\mathcal{F}_{fused}\) can be generated by weighted recalibration as
\begin{equation}
\mathcal{F}_{fused} = Conv_{1\times1}(\mathcal{F'}_{fine} \cdot W_{C2F} + \mathcal{F}_{coarse} \cdot (1 - W_{C2F})).
\end{equation}

Then, the fused feature \(\mathcal{F}_{fused}\) is processed with Layer Normalization (LN) and subsequently modified by the scale and shift factors \(\alpha_{2}\) and \(\beta_{2}\) as
\begin{equation}
\mathcal{F'}_{out} = LM(\alpha_{2} \odot \mathcal{F}_{fused} + \beta_{2}).
\end{equation}

Finally, 
the output feature of the C2FBlock, denoted as \(\mathcal{F}_{out}\), is obtained as 
\begin{equation}
\mathcal{F}_{out} = Conv_{1\times1}(SG(Conv_{1\times1}(\mathcal{F'}_{out}))).
\end{equation}

This multi-attentional feature complementation allows for highly effective integration of Coarse and Fine features, ensuring that both global context and fine details are preserved and enhanced. This results in a feature map that is not only contextually aware but also rich in local details, providing the foundation for superior image restoration performance.
\begin{table*}[t]
\centering
\caption{Comparison of image deraining results with SOTAs on the Rain100L deraining dataset~\cite{Rain100}.}
\label{Rain100L}
\renewcommand{\arraystretch}{1.1}
\resizebox{\textwidth}{!}{
\begin{tabular}{l|c|c|c|c|c|c|c}
\toprule[1.5pt]
\textbf{Methods}            & DerainNet~\cite{DerainNet} & SEMI~\cite{semi} & UMRL~\cite{UMRL} & DIDMDN~\cite{DIDMDN} & Jorder~\cite{jorder} & MSPFN~\cite{MSFSnet} & SPAIR~\cite{SPAIR} \\ 
\midrule
PSNR $\uparrow$             & 27.03 & 25.03 & 29.18 & 25.23 & 36.61 & 32.40 & 37.30 \\\hline
SSIM $\uparrow$             & 0.884 & 0.842 & 0.923 & 0.741 & 0.974 & 0.933 & 0.978 \\\hline
LPIPS $\downarrow$          & — & — & — & — & 0.028 & — & — \\ 
\midrule[1.5pt]
\textbf{Methods}            & MPRNet~\cite{Zamir2021MPRNet} & MAXIM~\cite{MAXIM}  & AirNet~\cite{Airnet} & IR-SDE~\cite{IRSDE} & MDARNet~\cite{MDARNet} & PromptIR~\cite{PromptIR} &M2PN~\cite{M2PN} \\ 
\midrule
PSNR $\uparrow$             & 36.40 & 38.06 & 34.90 & 38.30 & 35.68  & 37.04 & 38.36 \\\hline
SSIM $\uparrow$             & 0.965 & 0.977 & 0.966 & 0.981 & 0.961  & 0.979 & \firstone{0.985}\\\hline
LPIPS $\downarrow$          & 0.077 & 0.048 & — & \secondone{0.014}  & — & — & —\\ 
\midrule[1.5pt]
\textbf{Methods}   & NDR~\cite{NDR}        & GANet~
\cite{GANet}& FrePrompter~\cite{wu2025freprompter} &AnyIR~\cite{AnyIR} & VLU-Net~\cite{VLU} & Perceive-IR~\cite{perceive}  & \textbf{ReviveDiff (Ours)} \\ 
\midrule
PSNR $\uparrow$              & 38.33 & 37.53 & \secondone{38.75}&37.99 & 38.60 & 38.41  & \firstone{39.09} \\\hline
SSIM $\uparrow$          & \secondone{0.984}   & 0.973  & \secondone{0.984} &0.982& \secondone{0.984} & \secondone{0.984}  & 0.979 \\\hline
LPIPS $\downarrow$         & — & — & — & — & — & — & \firstone{0.012} \\ 
\bottomrule[1.5pt]
\end{tabular}}
\end{table*}

\begin{figure*}[!ht]
    \centering
    \captionsetup[subfigure]{labelformat=empty, justification=centering, font=footnotesize} 

    \setcounter{subfigure}{0}
    \subfloat[]{\includegraphics[width=0.14\textwidth]{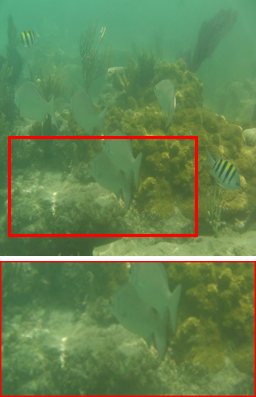}}\hfill
    \subfloat[]{\includegraphics[width=0.14\textwidth]{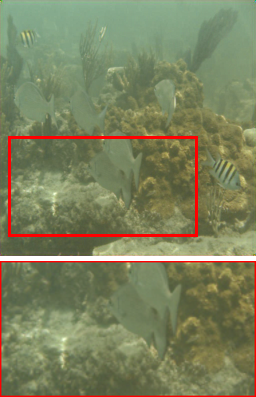}}\hfill
    \subfloat[]{\includegraphics[width=0.14\textwidth]{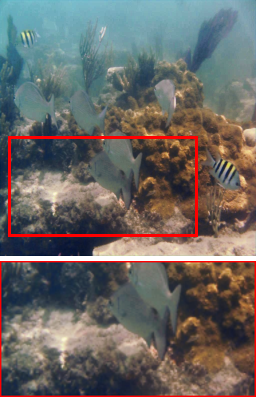}}\hfill
    \subfloat[]{\includegraphics[width=0.14\textwidth]{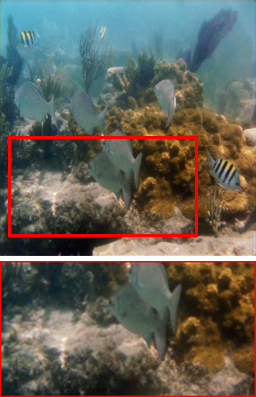}}\hfill
    \subfloat[]{\includegraphics[width=0.14\textwidth]{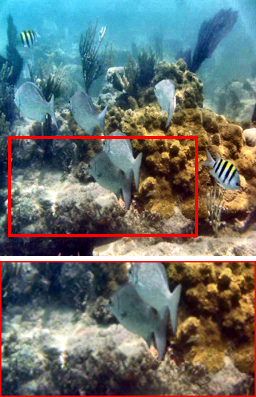}}\hfill
    \subfloat[]{\includegraphics[width=0.14\textwidth]{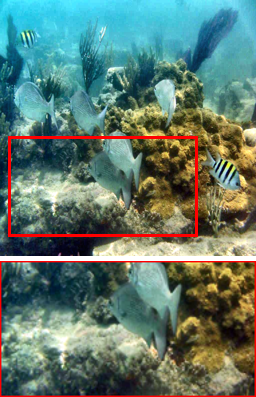}}\par\vspace{-1.3em}
    
    \setcounter{subfigure}{0}
    \subfloat[T90: Input]{\includegraphics[width=0.14\textwidth]{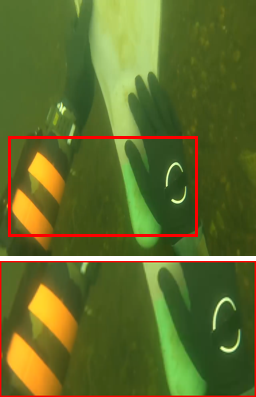}}\hfill
    \subfloat[UWCNN~\cite{uwcnn}]{\includegraphics[width=0.14\textwidth]{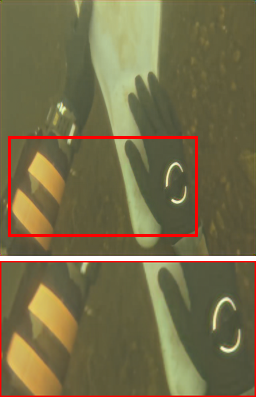}}\hfill
    \subfloat[FA+Net~\cite{fivenet}]{\includegraphics[width=0.14\textwidth]{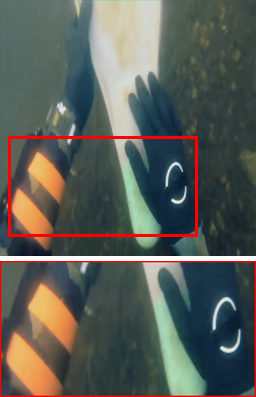}}\hfill
    \subfloat[NU2Net~\cite{NU2Net}]{\includegraphics[width=0.14\textwidth]{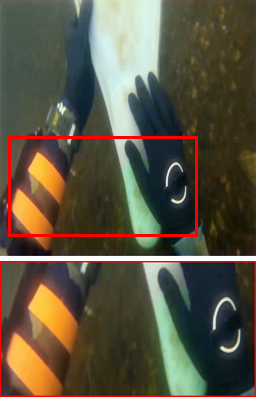}}\hfill
    \subfloat[\textbf{ReviveDiff (Ours)}]{\includegraphics[width=0.14\textwidth]{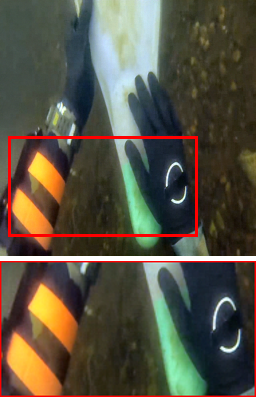}}\hfill
    \subfloat[GT]{\includegraphics[width=0.14\textwidth]{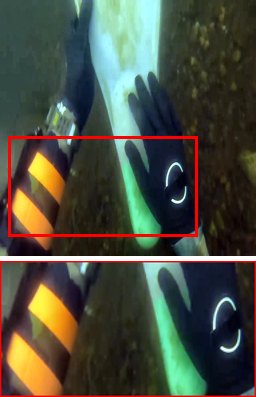}}\par\vspace{2mm}

    (a) With-Reference Results\par\medskip
    
    \setcounter{subfigure}{0}
    \subfloat[C60: Input]{\includegraphics[width=0.14\textwidth]{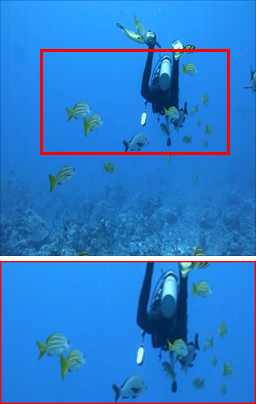}}\hfill
    \subfloat[UWCNN~\cite{uwcnn}]{\includegraphics[width=0.14\textwidth]{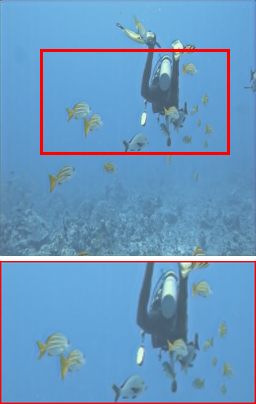}}\hfill
    \subfloat[FA+Net~\cite{fivenet}]{\includegraphics[width=0.14\textwidth]{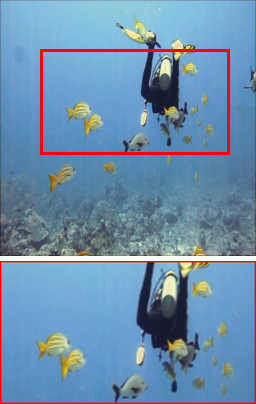}}\hfill
    \subfloat[NU2Net~\cite{NU2Net}]{\includegraphics[width=0.14\textwidth]{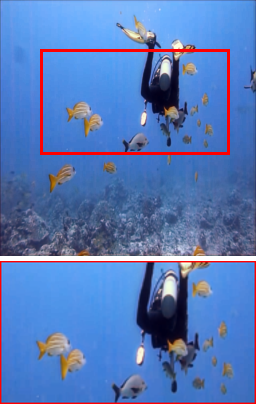}}\hfill
    \subfloat[UIEC2Net~\cite{uiec}]{\includegraphics[width=0.14\textwidth]{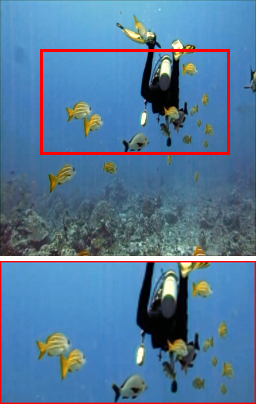}}\hfill
    \subfloat[\textbf{ReviveDiff (Ours)}]{\includegraphics[width=0.14\textwidth]{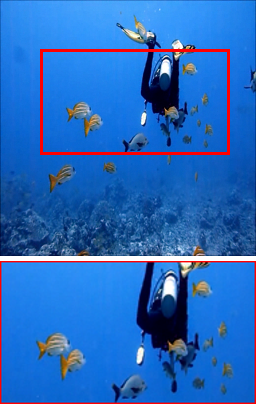}}\par\vspace{1mm}
    
    \setcounter{subfigure}{0}
    \subfloat[U45: Input]{\includegraphics[width=0.14\textwidth]{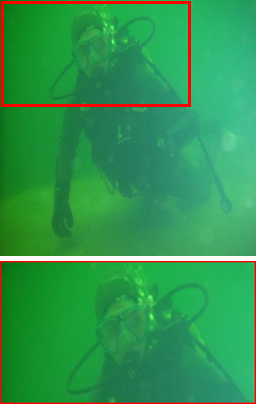}}\hfill
    \subfloat[UWCNN~\cite{uwcnn}]{\includegraphics[width=0.14\textwidth]{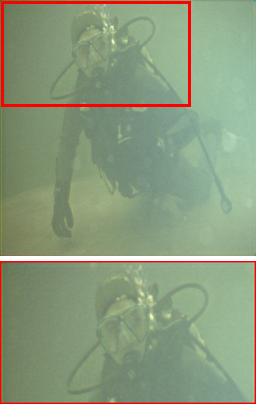}}\hfill
    \subfloat[FA+Net~\cite{fivenet}]{\includegraphics[width=0.14\textwidth]{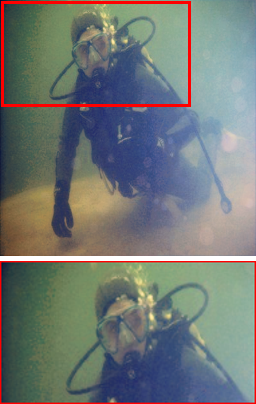}}\hfill
    \subfloat[NU2Net~\cite{NU2Net}]{\includegraphics[width=0.14\textwidth]{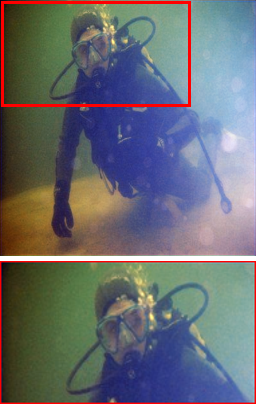}}\hfill
    \subfloat[UIEC2Net~\cite{uiec}]{\includegraphics[width=0.14\textwidth]{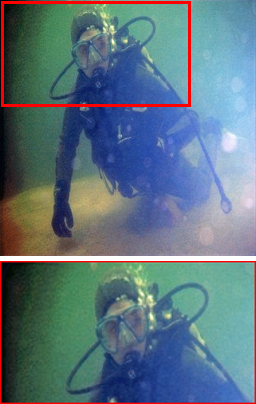}}\hfill
    \subfloat[\textbf{ReviveDiff (Ours)}]{\includegraphics[width=0.14\textwidth]{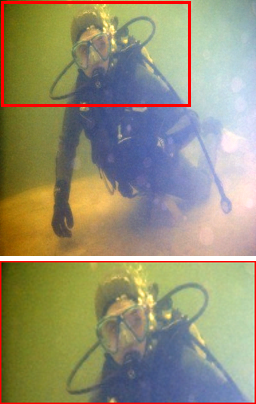}}\par\vspace{2mm}

    (b) Non-Reference Results\par\medskip
    \vspace{-0.5em}
    \caption{Visual comparison of underwater image enhancement results with SOTAs on T90~\cite{UIEB}, U45~\cite{U45}, and C60~\cite{UIEB}.}
    \label{uw}
\end{figure*}

\begin{table*}[h]
\caption{Quantitative comparison results of underwater image enhancement with SOTAs on the T90~\cite{UIEB}, U45~\cite{U45}, and C60~\cite{UIEB} underwater datasets.}
\label{underwater}
\resizebox{\textwidth}{!}{%
\begin{tabular}{lccccccccccc}
\toprule[1.5pt]
\multicolumn{1}{c|}{Dataset} &
  \multicolumn{1}{c|}{Methods} &
  \multicolumn{1}{c|}{GDCP~\cite{gdcp}} &
  \multicolumn{1}{c|}{UGAN~\cite{ugan}} &
  \multicolumn{1}{c|}{FUnIEGAN~\cite{funiegan}} &
  \multicolumn{1}{c|}{UWCNN~\cite{uwcnn}} &
  \multicolumn{1}{c|}{Ucolor~\cite{ucolor}} &
  \multicolumn{1}{c|}{MLLE~\cite{mlle}} &
  \multicolumn{1}{c|}{UShape~\cite{utrans}} &
  \multicolumn{1}{c|}{FA+Net~\cite{fivenet}} &
  \multicolumn{1}{c|}{NU2Net~\cite{NU2Net}} &
  \textbf{ReviveDiff} \\ \toprule[1.5pt]

\multicolumn{1}{c|}{\multirow{3}{*}{T90}} &
  \multicolumn{1}{c|}{PSNR $\uparrow$} &
  \multicolumn{1}{c|}{13.89} &
  \multicolumn{1}{c|}{17.42} &
  \multicolumn{1}{c|}{16.97} &
  \multicolumn{1}{c|}{19.02} &
  \multicolumn{1}{c|}{20.86} &
  \multicolumn{1}{c|}{19.48} &
  \multicolumn{1}{c|}{20.24} &
  \multicolumn{1}{c|}{20.98} &
  \multicolumn{1}{c|}{\secondone{22.93}} &
  \firstone{25.01} \\ \cline{2-12} 
\multicolumn{1}{c|}{} &
  \multicolumn{1}{c|}{SSIM $\uparrow$} &
  \multicolumn{1}{c|}{0.75} &
  \multicolumn{1}{c|}{0.76} &
  \multicolumn{1}{c|}{0.73} &
  \multicolumn{1}{c|}{0.82} &
  \multicolumn{1}{c|}{0.88} &
  \multicolumn{1}{c|}{0.84} &
  \multicolumn{1}{c|}{0.81} &
  \multicolumn{1}{c|}{0.88} &
  \multicolumn{1}{c|}{\secondone{0.90}} &
  \firstone{0.92} \\ \cline{2-12} 
\multicolumn{1}{c|}{} &
  \multicolumn{1}{c|}{NIQE $\downarrow$} &
  \multicolumn{1}{c|}{4.93} &
  \multicolumn{1}{c|}{5.81} &
  \multicolumn{1}{c|}{4.92} &
  \multicolumn{1}{c|}{4.72} &
  \multicolumn{1}{c|}{4.75} &
  \multicolumn{1}{c|}{4.87} &
  \multicolumn{1}{c|}{\secondone{4.67}} &
  \multicolumn{1}{c|}{4.83} &
  \multicolumn{1}{c|}{4.81} &
  \firstone{4.45} \\

\toprule[1.5pt]
\addlinespace[2mm]

\toprule[1.5pt]
\multicolumn{1}{c|}{\multirow{2}{*}{U45}} &
  \multicolumn{1}{c|}{UCIQE $\uparrow$} &
  \multicolumn{1}{c|}{\secondone{0.59}} &
  \multicolumn{1}{c|}{0.57} &
  \multicolumn{1}{c|}{0.56} &
  \multicolumn{1}{c|}{0.48} &
  \multicolumn{1}{c|}{0.58} &
  \multicolumn{1}{c|}{\secondone{0.59}} &
  \multicolumn{1}{c|}{0.57} &
  \multicolumn{1}{c|}{0.58} &
  \multicolumn{1}{c|}{0.57} &
  \firstone{0.62} \\ \cline{2-12} 
\multicolumn{1}{c|}{} &
  \multicolumn{1}{c|}{NIQE $\downarrow$} &
  \multicolumn{1}{c|}{4.14} &
  \multicolumn{1}{c|}{5.79} &
  \multicolumn{1}{c|}{4.45} &
  \multicolumn{1}{c|}{4.09} &
  \multicolumn{1}{c|}{4.71} &
  \multicolumn{1}{c|}{4.83} &
  \multicolumn{1}{c|}{4.20} &
  \multicolumn{1}{c|}{\secondone{4.02}} &
  \multicolumn{1}{c|}{5.7} &
  \firstone{3.93} \\

\toprule[1.5pt]
\addlinespace[2mm]

\toprule[1.5pt]
\multicolumn{1}{c|}{\multirow{2}{*}{C60}} &
  \multicolumn{1}{c|}{UCIQE $\uparrow$} &
  \multicolumn{1}{c|}{0.57} &
  \multicolumn{1}{c|}{0.55} &
  \multicolumn{1}{c|}{0.55} &
  \multicolumn{1}{c|}{0.51} &
  \multicolumn{1}{c|}{0.55} &
  \multicolumn{1}{c|}{0.57} &
  \multicolumn{1}{c|}{0.56} &
  \multicolumn{1}{c|}{0.57} &
  \multicolumn{1}{c|}{\secondone{0.58}} &
  \firstone{0.59} \\ \cline{2-12} 
\multicolumn{1}{c|}{} &
  \multicolumn{1}{c|}{NIQE $\downarrow$} &
  \multicolumn{1}{c|}{6.27} &
  \multicolumn{1}{c|}{6.90} &
  \multicolumn{1}{c|}{6.06} &
  \multicolumn{1}{c|}{5.94} &
  \multicolumn{1}{c|}{6.14} &
  \multicolumn{1}{c|}{5.85} &
  \multicolumn{1}{c|}{\secondone{5.60}} &
  \multicolumn{1}{c|}{5.70} &
  \multicolumn{1}{c|}{5.64} &
  \firstone{5.57} \\ \toprule[1.5pt]

\end{tabular}}
\end{table*}

\subsection{Prior-Guided Loss Functions}

In many related research fields, pixel-based loss functions such as L1, MSE, PSNR~\cite{PSNR}, and SSIM~\cite{SSIM} Losses are widely used. 
These loss functions aim to minimize pixel-wise differences between the generated image $I_{\text{gen}}$ and the ground truth image $I_{\text{gt}}$. 
The Pixel-based Loss, $L_{\text{pixel}}$, is defined as
\begin{equation}
L_{\text{pixel}}(I_{\text{gen}}, I_{\text{gt}}) = \frac{1}{N} \sum_{i=1}^{N} |I_{\text{gen},i} - I_{\text{gt},i}|,
\end{equation}
where $N$ denotes the total number of pixels in the image. 

However, for low-quality images captured from adverse natural conditions, visibility is predominantly affected by the medium of light propagation, such as fog or underwater environments, rather than by distortion or blur.  
As a result, objects within these images tend to retain their regular edges, shapes, and structures, even though overall visibility may be compromised. 
Leveraging this observation, incorporating edge and color information into the loss function can further enhance the training process by ensuring that key structural and color features are preserved during restoration and thus be advantageous.  


\subsubsection{Edge Loss}

In our study, to preserve structural information, we introduce an Edge Loss that focuses on maintaining the integrity of edges in the enhanced image. 
We utilize the Canny edge detector to extract the edge information from both the enhanced image $I_{\text{gen}}$ and the corresponding reference image $I_{\text{gt}}$.
The Edge Loss, $L_{\text{edge}}$, is defined as
\begin{equation}
L_{\text{edge}}(I_{\text{gen}}, I_{\text{gt}}) =  \left\| \mathcal{E}(I_{gen}; \sigma, T_{\text{low}}, T_{\text{high}}) - \mathcal{E}(I_{gt}; \sigma, T_{\text{low}}, T_{\text{high}}) \right\|_1.
\end{equation}
Here, $\mathcal{E}(\cdot; \sigma, T_{\text{low}}, T_{\text{high}})$ denotes the edge map obtained by applying the Canny edge detector with a Gaussian smoothing parameter $\sigma$ and hysteresis thresholding parameters 
$T_{\text{low}}$ and $T_{\text{high}}$. 
The Canny edge detection process, $\mathcal{E}$, includes gradient magnitude computation using a Gaussian filter with standard deviation $\sigma$, non-maximum suppression, and edge tracking by hysteresis using thresholds $T_{\text{low}}$ and $T_{\text{high}}$. 
The $\text{L}_1$ norm $ \left\| \cdot \right\|_1 $ calculates the sum of absolute differences between the edge maps of the enhanced and reference images, ensuring the preservation of edge information during image enhancement.

\begin{figure*}[h]
    \centering
    \captionsetup[subfigure]{labelformat=empty, font=footnotesize}
    \setcounter{subfigure}{0}

    \subfloat[Input]{\vtop{\null\hbox{\includegraphics[width=0.15\textwidth]{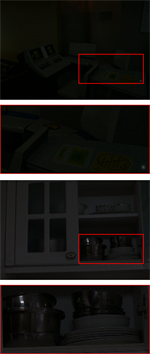}}}\vspace{-1mm}}\hfill
    \subfloat[LIME~\cite{LIME}]{\vtop{\null\hbox{\includegraphics[width=0.15\textwidth]{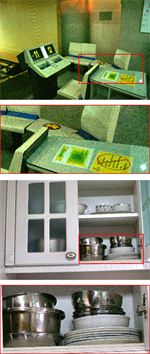}}}\vspace{-1mm}}\hfill
    \subfloat[NPE~\cite{NPE}]{\vtop{\null\hbox{\includegraphics[width=0.15\textwidth]{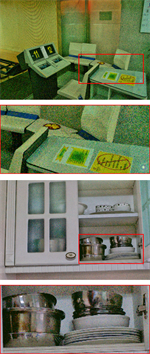}}}\vspace{-1mm}}\hfill
    \subfloat[DeepUPE~\cite{DeepUPE}]{\vtop{\null\hbox{\includegraphics[width=0.15\textwidth]{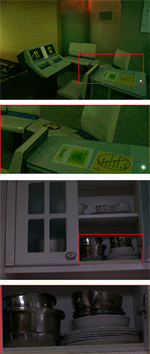}}}\vspace{-1mm}}\hfill
    \subfloat[GLAD~\cite{GLADNet}]{\vtop{\null\hbox{\includegraphics[width=0.15\textwidth]{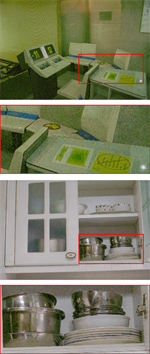}}}\vspace{-1mm}}\hfill
    \subfloat[Restormer~\cite{restormer}]{\vtop{\null\hbox{\includegraphics[width=0.15\textwidth]{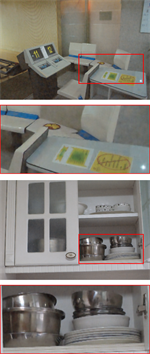}}}\vspace{-1mm}}\hfill
    \subfloat[EnlightenGAN~\cite{enlightengan}]{\vtop{\null\hbox{\includegraphics[width=0.15\textwidth]{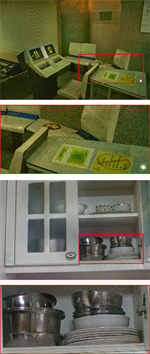}}}\vspace{-1mm}}\hfill
    \subfloat[Zero-DCE~\cite{zero}]{\vtop{\null\hbox{\includegraphics[width=0.15\textwidth]{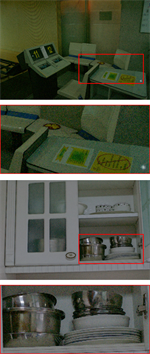}}}\vspace{-1mm}}\hfill
    \subfloat[RUAS~\cite{RUAS}]{\vtop{\null\hbox{\includegraphics[width=0.15\textwidth]{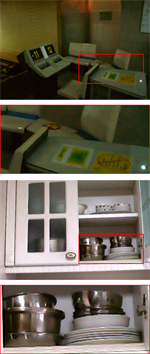}}}\vspace{-1mm}}\hfill
    \subfloat[LLFormer~\cite{LLformer}]{\vtop{\null\hbox{\includegraphics[width=0.15\textwidth]{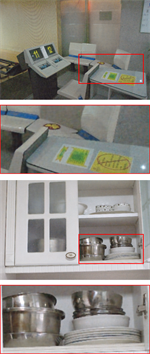}}}\vspace{-1mm}}\hfill
    \subfloat[\textbf{ReviveDiff (Ours)}]{\vtop{\null\hbox{\includegraphics[width=0.15\textwidth]{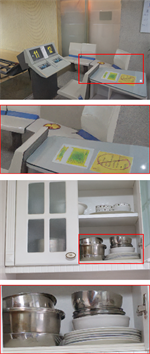}}}\vspace{-1mm}}\hfill
    \subfloat[GT]{\vtop{\null\hbox{\includegraphics[width=0.15\textwidth]{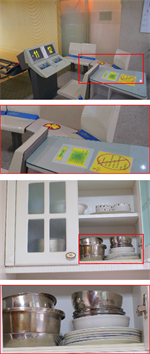}}}\vspace{-1mm}}\hfill

    \setcounter{subfigure}{0}

    \caption{Visual comparison of low-light image enhancement results with SOTAs on the LOL dataset~\cite{LOL_dataset}. The small
images enclosed in red below each result provide an enlarged view of the areas highlighted by red rectangles.} 
    \vspace{-0.5em}
    \label{fig:visual_comparsion_eval}
\end{figure*}

\subsubsection{Histogram Loss}

To ensure color consistency, we introduce a Histogram Loss to measure the discrepancy in color distribution between the enhanced image $I_{\text{gen}}$ and the corresponding reference image $I_{\text{gt}}$ in $R$, $G$, and $B$ channels,  respectively. 
The Histogram Loss, $L_{\text{hist}}$, 
can be defined as
\begin{equation}
L_{\text{hist}}(I_{\text{gen}}, I_{\text{gt}}) = \sum_{c=1}^{C} \left\| \frac{H(I_{\text{gen},c}, k)}{\sum_{k=1}^{K} H(I_{\text{gen},c}, k)} - \frac{H(I_{\text{gt},c}, k)}{\sum_{k=1}^{K} H(I_{\text{gt},c}, k)} \right\|_1,
\end{equation}
where the $H(I_{\text{gen},c}, k)$ and $H(I_{\text{gt},c}, k)$ are the histogram counts for the $k$-th bin in the $c$-th channel of the enhanced and reference images, respectively. 
$K$ denotes the total number of bins. 
The L1 norm $\left\| \cdot \right\|_1$ is used to calculate the sum of absolute differences between the normalized histograms of the enhanced and reference images across all bins and channels. 
This term ensures that the color distribution in the enhanced image closely matches that of the reference image, which is particularly important in challenging environments where color distortion is common.

\begin{table*}[t]
\caption{Quantitative comparisons of low-light image enhancement results on the LOL dataset~\cite{LOL_dataset}.} 
\label{lol}
\resizebox{\textwidth}{!}{
\begin{tabular}{l|c|c|c|c|c|c|c}
\toprule[1.5pt]
\textbf{Methods} & RetinexNet~\cite{retinexnet} & KIND~\cite{KIND} & DSLR~\cite{DPED} & DRBN~\cite{DRBN_CVPR} & Zero-DCE~\cite{zero} & MIRNet~\cite{recurisive} & EnlightenGAN~\cite{enlightengan} \\ \toprule[1.5 pt]
PSNR $\uparrow$ & 16.774 & 20.84 & 14.816 & 16.774 & 14.816 & \secondone{24.138} & 17.606 \\ \hline
SSIM $\uparrow$ & 0.462 & 0.790 & 0.572 & 0.462 & 0.572 & \secondone{0.830} & 0.653 \\\hline
LPIPS $\downarrow$ & 0.417 & 0.170 & 0.375 & 0.417 & 0.375 & 0.250 & 0.372 \\ 
\midrule[1.5pt]

\textbf{Methods} & RUAS~\cite{RUAS} & KIND++~\cite{KIND++} & DDIM~\cite{ddim} & CDEF~\cite{CDEF} & SCI~\cite{SCI} & URetinex-Net~\cite{retinexnet} & Uformer~\cite{Uformer} \\ \toprule[1.5 pt]
PSNR $\uparrow$ & 16.405 & 21.300 & 16.521 & 16.335 & 14.784 & 19.842 & 19.001 \\\hline
SSIM $\uparrow$ & 0.503 & 0.820 & 0.776 & 0.585 & 0.525 & 0.824 & 0.741 \\\hline
LPIPS $\downarrow$ & 0.364 & \secondone{0.160} & 0.376 & 0.407 & 0.366 & 0.237 & 0.354 \\ 
\midrule[1.5pt]

\textbf{Methods} & Restormer~\cite{restormer}  & UHDFour~\cite{UHDFour} & WeatherDiff~\cite{weather} & LLformer~\cite{LLformer} & LightenDiffusion~\cite{LightenDiffusion} &PromptGIP~\cite{PROMPTGIP}& \textbf{ReviveDiff (Ours)} \\ \toprule[1.5 pt]
PSNR $\uparrow$ & 20.614   & 23.093 & 17.913 & 23.650 & 20.453&20.30 & \firstone{24.272} \\\hline
SSIM $\uparrow$ & 0.797 & 0.821 & 0.811 & 0.816 & 0.803 &0.803& \firstone{0.832} \\\hline
LPIPS $\downarrow$ & 0.288 & 0.259 & 0.272 & 0.171 & 0.192&N/A & \firstone{0.0875} \\ 
\bottomrule[1.5pt]
\end{tabular}}
\end{table*}

\subsubsection{Combined Loss}

The overall loss function used to guide the training of our model combines the pixel-wise, edge, and histogram losses to ensure that the restored image maintains not only pixel-level accuracy but also structural integrity and accurate color representation. 

The combined loss, $L$, is formulated as 
\begin{equation}
L = \lambda_{\text{1}} L_{\text{pixel}}(I_{\text{gen}}, I_{\text{gt}}) + \lambda_{\text{2}} L_{\text{edge}}(I_{\text{gen}}, I_{\text{gt}}) + \lambda_{\text{3}} L_{\text{hist}}(I_{\text{gen}}, I_{\text{gt}}).
\end{equation}
Here, $\lambda_{\text{1}}$, $\lambda_{\text{2}}$, and $\lambda_{\text{3}}$ are the learned weighting coefficients that dynamically balance the contributions of the Pixel-wise loss, Edge loss, and Histogram loss, respectively. 

These weights allow the model to focus on different aspects of image quality—pixel-level accuracy, edge preservation, and color consistency—based on the specific degradation present in the input image.


\section{Experiments}
\label{sec:Experiments}	

To demonstrate the superior performance of our proposed ReviveDiff approach in enhancing image visibility under adverse conditions, we designed comprehensive experiments that benchmark our approach with SOTA approaches on five challenging tasks: image deraining, low-light image enhancement, nighttime dehazing, desmoking, and underwater image enhancement. 
Next, we first introduce the datasets and experiment details. Then, we present experimental comparisons with SOTA approaches and ablation studies on the key modules in our approach.


\subsection{Datasets} 

\textbf{Image Deraining:}
For the image deraining task, we evaluated our ReviveDiff model using the Rain100L~\cite{Rain100} dataset, which includes both clean and synthetic rainy images, with 200 pairs for training and 100 for testing.\\
\textbf{Underwater Image Enhancement:}
For the underwater image enhancement task, we used three datasets: the UIEB dataset~\cite{UIEB}, the C60~\cite{UIEB} dataset, and the U45 dataset~\cite{U45}. 
The UIEB dataset consists of 890 pairs of low-quality and high-quality images, while the C60 and U45 datasets comprise 60 and 45 challenging images, respectively, without reference images. 
Consistent with previous works, we split the UIEB dataset into 800 pairs for training and 90 images for testing (referred to as ``T90''). We tested our ReviveDiff on all three datasets.\\
\textbf{Low-light Image Enhancement:}
For the low-light image enhancement task, we evaluated our ReviveDiff on the LOw-Light (LOL)~\cite{LOL_dataset} dataset, which comprises 500 image pairs of synthetic low-light and normal-light images. The LOL dataset provides 485 pairs for training and 15 pairs for testing.\\
\textbf{Nighttime Dehazing:}
For the nighttime dehazing task, we used the GTA5~\cite{gta5} dataset, a synthetic dataset created with the GTA5 game engine. The GTA5 dataset includes 864 image pairs, with 787 of them designated for training and the remaining 77 pairs for testing.\\
\textbf{Image Desmoking:}
For the image desmoking task, we utilized the SMOKE~\cite{SMOKE} dataset, which consists of 132 paired real-world images captured in natural scenes with fog generated by a fog machine. 
120 pairs of these images are used for training, and the remaining 12 pairs for testing.

\begin{figure*}[ht]
	\centering
	\captionsetup[subfigure]{labelformat=empty,font=footnotesize}
        \setcounter{subfigure}{0}
	\subfloat[Input]{\includegraphics[width = 0.196\textwidth]{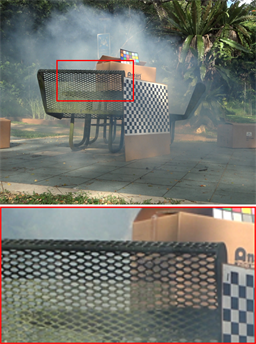}\vspace{-1mm}}\hfill
	\subfloat[DeHamer~\cite{guo2022image}]{\includegraphics[width = 0.196\textwidth]{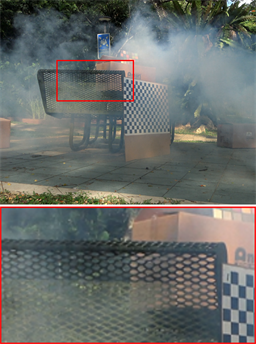}\vspace{-1mm}}\hfill
	\subfloat[DehazeFormer~\cite{song2023vision}]{\includegraphics[width = 0.196\textwidth]{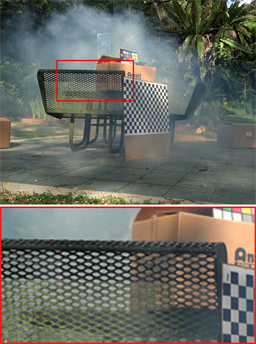}\vspace{-1mm}}\hfill
	\subfloat[D4~\cite{yang2022self}]{\includegraphics[width = 0.196\textwidth]{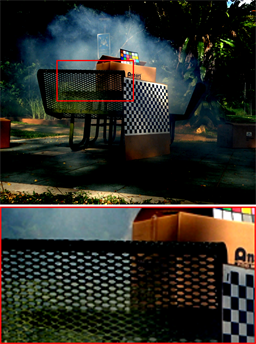}\vspace{-1mm}}\hfill
	\subfloat[PSD~\cite{chen2021psd}]{\includegraphics[width = 0.196\textwidth]{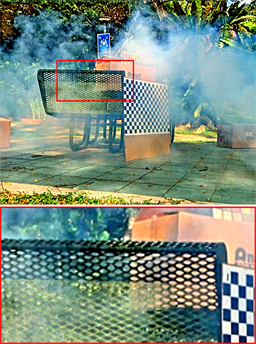}\vspace{-1mm}}\hfill
	\subfloat[4K~\cite{zheng2021ultra}]{\includegraphics[width = 0.196\textwidth]{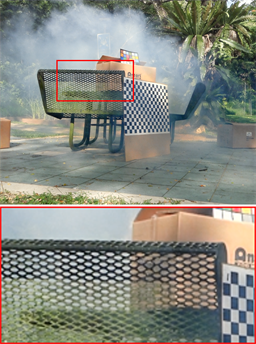}\vspace{-1mm}}\hfill
	\subfloat[MSBDN~\cite{dong2020multi}\vspace{-1mm}]{\includegraphics[width = 0.196\textwidth]{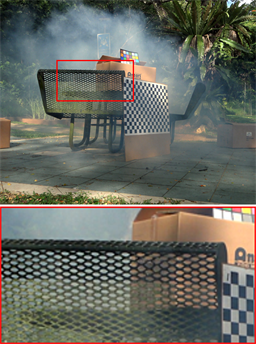}\vspace{-1mm}}\hfill
	\subfloat[DAN~\cite{shao2020domain}]{\includegraphics[width = 0.196\textwidth]{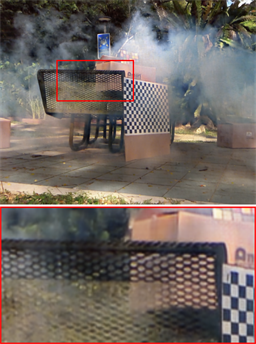}\vspace{-1mm}}\hfill
	\subfloat[SMOKE~\cite{SMOKE}]{\includegraphics[width = 0.196\textwidth]{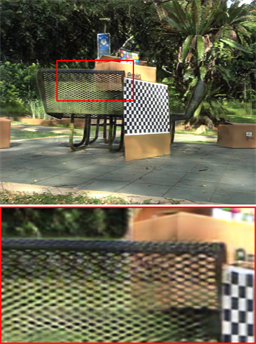}\vspace{-1mm}}\hfill
 	\subfloat[\textbf{ReviveDiff (Ours)}]{\includegraphics[width = 0.196\textwidth]{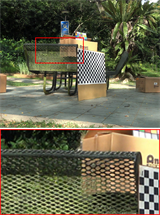}\vspace{-1mm}}\hfill
    \caption{ Visual comparison of smoke image enhancement results with SOTAs on the SMOKE dataset~\cite{SMOKE}.}
    \vspace{-1em}
    \label{SMOKE:fig}
\end{figure*}

\subsection{Implementation Details}

During model training, we utilized a batch size of 4 and initialized the learning rate to $1 \times 10^{-4}$. 
We used the Adam optimizer~\cite{adam}, with parameters $\beta_{1}$ and $\beta_{2}$ to 0.9 and 0.999, for training optimization. 
The learning rate was adjusted using the MultiStepLR strategy for decay. The total number of training iterations varied based on the specific degradation being addressed. 
We employed a noise level of \( \sigma = 90 \) and set the diffusion step to 300 for all tasks. All experiments were conducted on a single NVIDIA RTX 4090 GPU using the PyTorch platform.

\subsection{Evaluation Metrics}
In our experiments, we adopt several widely used evaluation metrics to assess our model's performance across various restoration tasks comprehensively. We employ the Peak Signal-to-Noise Ratio (PSNR)~\cite{PSNR} and the Structural Similarity Index Measure (SSIM)~\cite{SSIM} to quantify the fidelity of the restored images compared to the ground truth. Additionally, the Learned Perceptual Image Patch Similarity (LPIPS)~\cite{LPIPS} metric is used to evaluate perceptual similarity, reflecting human visual perception. For tasks such as underwater image enhancement, where reference images may not be available, we further utilize no-reference metrics like the Natural Image Quality Evaluator (NIQE)~\cite{NIQE} and the Underwater Color Image Quality Evaluation (UCIQE)~\cite{UCIQE} to assess naturalness and color accuracy. Together, these metrics provide a robust framework for evaluating both pixel-level accuracy and perceptual quality across diverse adverse conditions.

\subsection{Comparisons with the State of the Arts}
\label{sec:exp:compare}

\subsubsection{Image Deraining}

Images captured in rainy conditions often suffer from the detrimental effects of rain streaks. 
To evaluate the deraining capability of our method, we utilized three widely used metrics: PSNR, SSIM, and LPIPS.
We compared our method with several SOTAs from 2016 to 2025.

As shown in Table~\ref{Rain100L}, our ReviveDiff model achieves a competitive performance across all these metrics. 
When compared to the second-best method, FrePrompter~\cite{wu2025freprompter}, our PSNR surpassed it by 0.34, making ReviveDiff the only method to achieve a PSNR over 39. 
This indicates the superior quantitative performance of our ReviveDiff in deraining. 
Visually, as depicted in Fig.~\ref{Rain100L:fig}, our results show a significant reduction in rain presence and are most closely aligned with the reference image in terms of quality.

\subsubsection{Underwater Image Enhancement}

Underwater images often suffer from degradation due to the scattering and absorption effects of water, coupled with low illumination, resulting in poor visual quality. In our study, we evaluated ReviveDiff using both paired (T90) and non-reference datasets (U45 and C60). For the T90 dataset, we employed widely recognized metrics: PSNR and SSIM, to assess image quality. 
Additionally, we utilized two metrics that do not require reference images for computation: UCIQE (to evaluate brightness) and NIQE (to assess distortion and noise). These metrics were also applied to the U45 and C60 datasets. 
We compared our ReviveDiff model with SOTAs~\cite{uiec}, among which GDCP~\cite{gdcp} represents traditional theory-based approaches, while the others are deep learning-based. 

For a quantitative comparison, as shown in Table~\ref{underwater}, our ReviveDiff outperforms all other methods across all three datasets. 
Specifically, ReviveDiff achieves exceptional performance in PSNR and SSIM metrics. Compared to the second-best method, NU2Net~\cite{NU2Net}, our PSNR is higher by 2.08, and ReviveDiff is the only method to achieve an SSIM greater than 0.9. 
Furthermore, on the U45 dataset, the proposed ReviveDiff is the only method to achieve an NIQE score lower than 4, indicating its superior restoration performance of brightness. 

For visual quality comparison, as illustrated in Fig.~\ref{uw}, our ReviveDiff generates results that are most similar to the reference images in the T90 dataset. Additionally, for the non-reference datasets U45 and C60, our results also demonstrate better visual quality.

\begin{table}[t]
\centering
\caption{Comparison of image desmoking results on the SMOKE dataset~\cite{SMOKE}. } 
\renewcommand{\arraystretch}{1.2}
\setlength{\tabcolsep}{15pt}
\begin{tabular}{l|c|c}
\toprule[1.2pt]
Methods & PSNR $\uparrow$ & SSIM $\uparrow$ \\ \toprule[1.5 pt]
DCP~\cite{dcp}             & 11.26 & 0.26 \\ \hline
GDN~\cite{gdn}             & 15.19 & 0.53 \\ \hline
MSBDN~\cite{dong2020multi} & 13.19 & 0.34 \\  \hline
DeHamer~\cite{guo2022image}& 13.31 & 0.28 \\ \hline
SMOKE~\cite{SMOKE}         & 18.83 & 
0.62 \\ \hline
CTHD-Net~\cite{li2024cthd}& \secondone{19.23} & 
\secondone{0.63} \\ \hline
\textbf{ReviveDiff (Ours)}              & \firstone{20.09} & \firstone{0.65} \\ 
\bottomrule[1.2pt]
\end{tabular}
\label{SMOKE}
\end{table}

\begin{table}[t]
\caption{Ablation studies on the UIEB underwater dataset~\cite{UIEB}.} 
\centering
\label{Underwater}
\begin{tabular}{l|c|c|c}
\toprule[1.5 pt]
Methods   & PSNR $\uparrow$           & SSIM $\uparrow$            & NIQE $\downarrow$  \\ \toprule[1.5 pt]
Only $L_1$ Loss & 24.47    & \secondone{0.91}          & 4.61   \\ \hline
$w/o$ Edge Prior  & 24.26    & \firstone{0.92}          & 4.51   \\ \hline

$w/o$ His Prior  & 24.25    & \firstone{0.92}         & \secondone{4.48}    \\ \hline
$w/o$ MAFC  & 20.08    & 0.86          & 4.65   \\ \hline
$w/o$ C2F  & 24.04    & \firstone{0.92}          & 4.51   \\ \hline
$W_{1}:W_{2}$ = 1:1 & \secondone{24.64}    & \firstone{0.92}          & 4.58   \\ \hline
$W_{1}:W_{2}$ = 10:1 & 24.55    & \firstone{0.92}          & 4.56   \\ \hline
$W_{1}:W_{2}$ = 1:10 & 23.58    & \secondone{0.91}        & 4.53   \\ \hline
\textbf{ReviveDiff (Ours)}   & \firstone{25.01} & \firstone{0.92}    & \firstone{4.45}\\ \toprule[1.5 pt]
\end{tabular}
\end{table}

\subsubsection{Low-light Image Enhancement}

Enhancing the image quality of low-light images is a significant challenge in current research. To assess the performance of our ReviveDiff model, we employed metrics such as PSNR, SSIM, and LPIPS and compared our results with SOTA methods from 2018 to 2024. 

As shown in Table~\ref{lol}, our ReviveDiff model outperforms competing approaches across these metrics. Notably, our method is the only one to achieve an LPIPS score lower than 0.1, signifying unparalleled results in terms of human visual perception. Furthermore, as depicted in Fig.~\ref{fig:visual_comparsion_eval}, our results demonstrate superior visual quality, exhibiting minimal color distortion and optimal brightness, closely matching the ground truth images.

\subsubsection{Nighttime Image Dehazing}

\begin{figure*}[t]
	\centering
	\includegraphics[width = 1\linewidth]{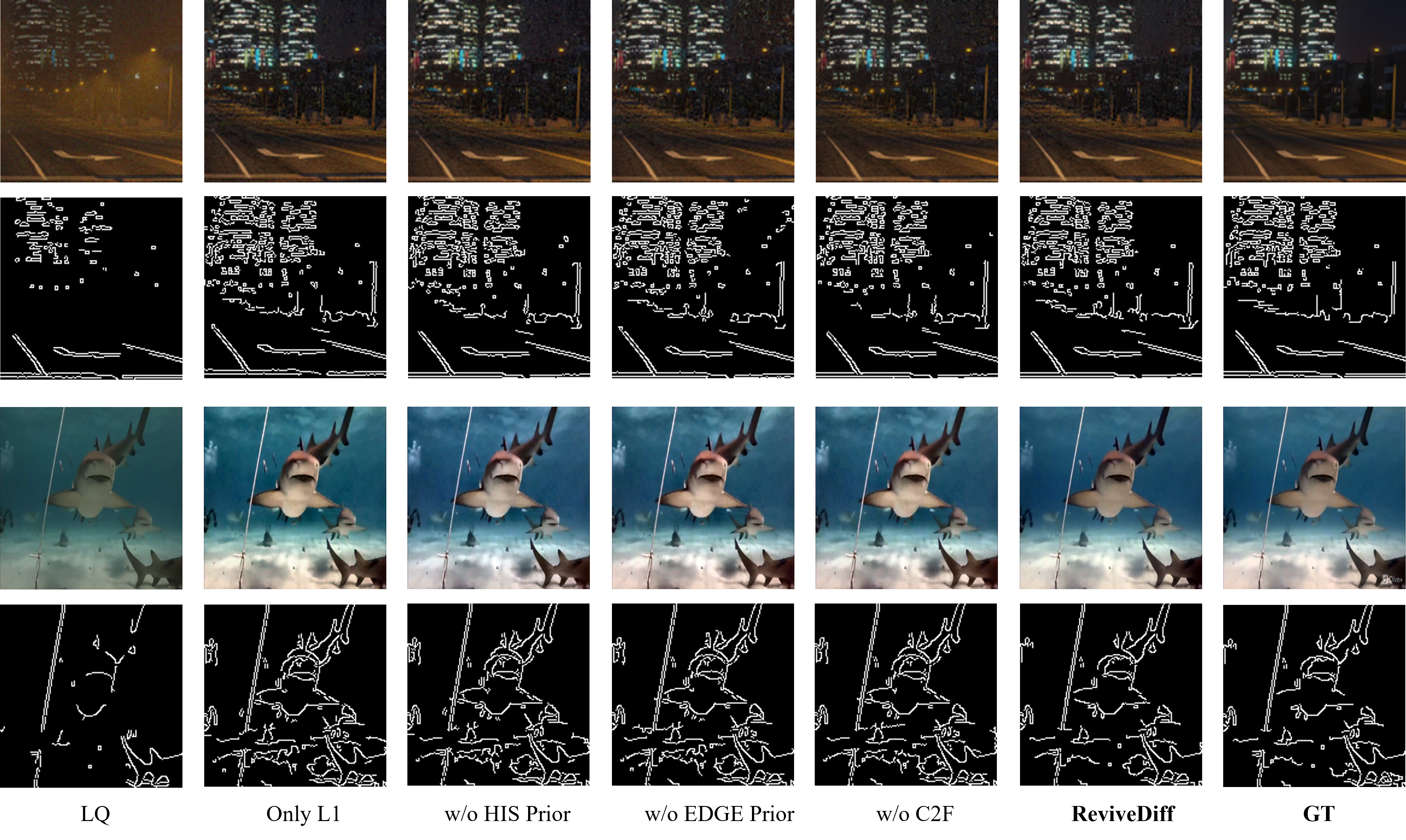}
 \vspace{-1em}
        \caption{Visual comparison of the ablation study results and edge maps on the GTA5 nighttime hazing dataset~\cite{gta5} and the UIEB underwater image dataset~\cite{UIEB}.}
	\label{ablationfig}
\end{figure*}

Nighttime image Dehazing presents a highly complex task, requiring methods that preserve the low-light environment while simultaneously removing haze. As demonstrated in Table~\ref{GTA5}, our ReviveDiff model surpasses SOTAs in both PSNR and LPIPS metrics, experiencing a marginal loss in SSIM by only 0.009 (1\%) compared to the method by Jin \textit{et al.}~\cite{gta5}. However, our ReviveDiff model outperforms that of Jin \textit{et al.} in PSNR by 2.53 (8\%), and a lower LPIPS score indicates better alignment with human perception.  This is further supported by visual quality comparisons in Fig.~\ref{ablationfig}, where it is evident that the method by Jin \textit{et al.}~\cite{gta5} fails to restore light as effectively as ours, resulting in darker images with lower contrast.


\subsubsection{Real-world Image Desmoking}

Real-world image desmoking is a challenging task that requires methods capable of effectively removing dense fog while preserving the underlying image details and maintaining natural visibility.

Table~\ref{SMOKE} shows a quantitative comparison of desmoking performance. As shown in this table, our method achieves the highest PSNR of 20.09 dB and SSIM of 0.65, outperforming other SOTA methods such as SMOKE~\cite{SMOKE} (PSNR 18.83 dB, SSIM 0.62) and CTHD-Net~\cite{li2024cthd} (PSNR 19.23 dB, SSIM 0.63). This significant improvement highlights the superior capability of our method in restoring clear images while preserving structural information. The results demonstrate that our approach surpasses existing techniques, achieving state-of-the-art performance in the image-desmoking task.

\subsection{Ablation Studies}
    
We performed additional ablation studies to validate the effectiveness of our key components: Coarse-to-Fine Learning (C2F), Multi-Attentional Feature Complementation (MAFC), and the guided losses based on Edge and Histogram Priors. Quantitative comparisons on the UIEB underwater dataset and the GTA5 nighttime hazing dataset are presented in Tables~\ref{Underwater} and~\ref{GTA5}, respectively. Fig.~\ref{ablationfig} provides visual comparisons, including edge maps, highlighting the influence of each ablated component in these challenging scenarios.
\subsubsection{Effectiveness of Coarse-to-Fine Learning} Tables~\ref{Underwater} and~\ref{GTA5} show that even without Coarse-to-Fine Learning (denoted as $w/o\ C2F$), our model still outperforms some competing methods. However, it falls short of the full ReviveDiff model in terms of PSNR, SSIM, LPIPS, and NIQE, confirming the importance of progressively capturing both coarse and fine features for robust enhancement in nighttime dehazing and underwater tasks. Moreover, as the edge maps illustrate, omitting Coarse-to-Fine Learning leads to significant detail loss in nighttime dehazing images in underwater images. For instance, in the visualization of the underwater image, the result of $w/o\ C2F$ exhibits inconsistent colors. Moreover, its edge map of the nighttime defogging image loses some structural details; conversely, the edge map of the underwater image reinforces unnecessary details.




\begin{table}[!t]
\caption{Ablation studies on the GTA5 nighttime dehazing dataset~\cite{gta5}.} 
\centering
\label{GTA5}
\begin{tabular}{l|c|c|c}
\toprule[1.5 pt]
Methods   & PSNR $\uparrow$           & SSIM $\uparrow$            & LPIPS $\downarrow$         \\ \toprule[1.5 pt]
Zhang \textit{et al.}~\cite{zhang2017fast}    & 20.92         & 0.646          & -              \\ \hline
Ancuti \textit{et al.}~\cite{ancuti2016night}   & 20.59         & 0.623          & -              \\ \hline
Yan \textit{et al.}~\cite{yan2020nighttime}      & 27.00         & 0.850          & -              \\ \hline
CycleGAN~\cite{zhu2017unpaired} & 21.75         & 0.696           & -              \\ \hline
Jin \textit{et al.}~\cite{gta5}       & 30.38         & \firstone{0.904} & 0.099          \\ \hline

Only $L_1$ Loss & 30.82    & 0.840          & 0.103    \\ \hline
$w/o$ Edge Prior  & 30.56    & 0.838          & 0.107    \\ \hline

$w/o$ His Prior  & 30.49    & 0.837          & 0.106    \\ \hline
$w/o$ MAFC  & 23.77    & 0.722          & 0.206    \\ \hline
$w/o$ C2F  & \secondone{32.52}    & 0.840          & \secondone{0.098}    \\ \hline
\textbf{ReviveDiff (Ours)}   & \firstone{32.91} & \secondone{0.895}    & \firstone{0.094} \\ \toprule[1.5 pt]
\end{tabular}
 \vspace{-1em}
\end{table}

\subsubsection{Effectiveness of Multi-attentional Feature Complementation} 
The variant without MAFC (denoted as $w/o\ MAFC$), which replaces MAFC with a simple addition operation, suffers a significant performance drop in both Table~\ref{Underwater} and Table~\ref{GTA5}. 
For instance, on the UIEB dataset (Table~\ref{Underwater}), removing MAFC reduces PSNR from 25.01\,dB to 20.08\,dB and SSIM from 0.92 to 0.86, indicating poorer overall restoration. On the GTA5 dataset (Table~\ref{GTA5}), it similarly lowers performance by around 0.4\,dB in PSNR and increases LPIPS by over 0.01 compared to the full ReviveDiff model. 
These declines highlight MAFC’s critical 
role in balancing and fusing coarse and fine features effectively.

\begin{table}[!t]
\centering
\caption{Efficiency comparisons between similar SOTAs and our ReviveDiff, the parameters and Multi-Adds are computed with an input size of 3$\times$256$\times$256.}
\resizebox{\columnwidth}{!}{
\begin{tabular}{l|c|c|c|c}
\toprule[1.6pt]
\textbf{Dataset} & \textbf{Methods} & \textbf{Parameters} & \textbf{Multi-Adds} & \textbf{PSNR/SSIM/LPIPS(NIQE)} \\ \toprule[1.5 pt]
\multirow{3}{*}{Rain100L~\cite{Rain100}} & IR-SDE~\cite{IRSDE} & 137.15M & 379.31G & 38.30 / \textbf{0.981} / 0.014 \\
 & AWRaCLe~\cite{AWRaCLe} & 94.25M & 158.64G & 35.71 / 0.966 / - \\
 & \textbf{ReviveDiff (Ours)} & \textbf{88.31M} & \textbf{73.95G} & \textbf{39.09} / \underline{0.979} / \textbf{0.012} \\
\midrule
\multirow{2}{*}{Nighttime~\cite{gta5}} & IR-SDE~\cite{IRSDE} & 137.15M & 379.31G & 31.34 / 0.856 / 0.098 \\
 & \textbf{ReviveDiff (Ours)} & \textbf{88.31M} & \textbf{73.95G} & \textbf{32.91 / 0.895 / 0.094} \\
 \midrule
 \multirow{2}{*}{UIEB~\cite{UIEB}} & IR-SDE~\cite{IRSDE} & 137.15M & 379.31G & 21.55 / 0.89 / 4.75\\
 & \textbf{ReviveDiff (Ours)} & \textbf{88.31M} & \textbf{73.95G} & \textbf{25.01 / 0.92 / 4.45} \\
\midrule
\multirow{2}{*}{LOL~\cite{LOL_dataset}} & WeatherDiff~\cite{weather} & 87.90M & 261.55G & 17.913 / 0.811 / 0.272 \\
 & \textbf{ReviveDiff (Ours)} & \textbf{88.31M} & \textbf{73.95G} & \textbf{24.272 / 0.832 / 0.0875} \\

\bottomrule[1.5pt]
\end{tabular}}
\label{multi}
\end{table}

\subsubsection{Discussion on Dynamic Weighting in MAFC}
To further validate the effectiveness of the proposed dynamic weighting strategy within the MAFC module, we conducted an additional ablation study on the UIEB dataset by comparing three fixed weight ratios ($W_{1}:W_{2}$) against our adaptive weighting approach. Specifically, we tested: (1) $W_{1}:W_{2}=1:1$, (2) $W_{1}:W_{2}=10:1$, and (3) $W_{1}:W_{2}=1:10$. As shown in Table~\ref{Underwater}, all fixed-weight variants exhibit noticeable performance degradation compared to our dynamic weighting strategy. For instance, the equal-weight assignment ($1:1$) achieves a PSNR of 24.64\,dB and a NIQE of 4.58, whereas the highly skewed ratio ($1:10$) further degrades the PSNR to 23.58\,dB and SSIM to 0.91. In contrast, our adaptive weighting mechanism reaches a superior PSNR of 25.01\,dB and a lower NIQE of 4.45. These quantitative results explicitly validate that dynamically adjusting the weighting factors between coarse and fine features significantly enhances performance, confirming the critical role of adaptive weighting in addressing diverse degradations present in underwater images.

\subsubsection{Effectiveness of the Prior-Guided Losses}
Fig.~\ref{ablationfig} and Tables~\ref {Underwater} and~\ref{GTA5} present the results for $w/o\ Edge\ Prior$ and $w/o\ His\ Prior$, demonstrating that omitting these priors degrades both boundary preservation and color fidelity. In particular, the $Only\ L_1\ Loss$ variant shows that relying solely on a pixel-based metric is insufficient for handling complex degradations, often yielding unnatural color distributions and less-defined edges. As seen in Fig.~\ref{ablationfig}, $w/o\ Edge\ Prior$ fails to maintain sharp 
object boundaries. At the same time, $w/o\ His\ Prior$ leads to noticeable color inconsistencies—especially in underwater images, where the restoration process suffers from an uncorrected color cast. In contrast, incorporating both Edge and Histogram priors simultaneously sharpens edges and refines color accuracy. The improved edge maps in the lower rows of Fig.~\ref{ablationfig} verify that these guided losses help preserve subtle details and rectify color distortions under challenging low-visibility conditions. This highlights how the proposed prior-guided losses address the limitations of purely pixel-based objectives, ultimately enabling more robust restoration in adverse environments.

\subsubsection{Efficiency Analysis} Table~\ref{multi} provides a comprehensive efficiency comparison of our proposed ReviveDiff model against similar state-of-the-art methods, including IR-SDE~\cite{IRSDE}, WeatherDiff~\cite{weather}, and AWRaCLe~\cite{AWRaCLe}, in terms of model complexity (number of parameters), computational cost (Multi-Adds), and restoration performance across multiple datasets. Our proposed ReviveDiff significantly reduces computational overhead, requiring only 88.31M parameters and 73.95G Multi-Adds. Despite this reduced complexity, ReviveDiff consistently achieves superior performance compared to other methods, notably surpassing IR-SDE on Rain100L (39.09 dB vs. 38.30 dB PSNR), Nighttime dehazing (32.91 dB vs. 31.34 dB PSNR), and UIEB underwater datasets (25.01 dB vs. 21.55 dB PSNR). Additionally, ReviveDiff significantly outperforms WeatherDiff on the LOL dataset (24.272 dB vs. 17.913 dB PSNR) and AWRaCLe on Rain100L (39.09 dB vs. 35.71 dB PSNR). These results underscore ReviveDiff's efficiency and effectiveness, demonstrating its suitability for practical deployment in resource-constrained environments.


\section{Conclusion}
\label{sec:Conclusion}

In this paper, we developed a novel diffusion model, “ReviveDiff”, specifically designed for enhancing images degraded by diverse adverse environmental conditions such as fog, rain, underwater scenarios, and nighttime settings. ReviveDiff features a unique Coarse-to-Fine Learning framework and a Multi-Attention Feature Complementation module, enabling the effective learning and fusion of multi-scale features. Additionally, it incorporates edge and histogram priors for enhanced structure and color restoration. Upon evaluation across seven datasets covering a range of adverse degradations, ReviveDiff demonstrably outperforms existing SOTA methods, offering a robust solution for universal adverse condition image enhancement challenges. To our knowledge, ReviveDiff represents a breakthrough contribution by utilizing a diffusion-based universal approach to address adverse conditions. This work not only underscores the versatility and efficacy of diffusion models in restoring images affected by adverse conditions but also establishes a new benchmark for future research. Looking ahead, we plan to explore strategies for reducing the number of diffusion steps and accelerating inference, further enhancing the model’s efficiency without compromising its restoration quality.


\bibliographystyle{IEEEtran} 
\bibliography{sample-base}

\end{document}